\pdfoutput=1

\documentclass[11pt]{article}

\usepackage[final]{acl}

\usepackage{times}
\usepackage{latexsym}
\usepackage{hyperref}       
\usepackage{algorithm}
\usepackage{mleftright}
\usepackage{algpseudocode}
\definecolor{deepblue}{RGB}{0,0,139}
\hypersetup{citecolor=deepblue}
\usepackage{url}            
\usepackage{booktabs}       
\usepackage{amsfonts}       
\usepackage{nicefrac}       
\usepackage{microtype}      
\usepackage{xcolor}         
\usepackage{amsmath}
\usepackage{amssymb}
\usepackage{mathtools}
\usepackage{amsthm}
\usepackage{cancel}
\usepackage{subcaption}
\usepackage{dsfont}

\usepackage[flushmargin,hang]{footmisc}
\usepackage[inline]{enumitem}
\setitemize{noitemsep,topsep=0pt,parsep=0pt,partopsep=0pt,leftmargin=*}
\setenumerate{noitemsep,topsep=0pt,parsep=0pt,partopsep=0pt,leftmargin=*}

\usepackage[T1]{fontenc}

\usepackage[utf8]{inputenc}

\usepackage{float}

\usepackage{linguex}
\usepackage{CJKutf8}

\usepackage{inconsolata}

\usepackage{graphicx}

\usepackage{tikz}
\usetikzlibrary{automata, positioning, calc, arrows.meta}
\newcommand{\emojiimg}[1]{\raisebox{-0.2ex}{\includegraphics[height=1em]{#1}}}

\usepackage{rycolab}
\definecolor{ETHPetrol}{RGB}{0,120,148}	
\colorlet{MacroColor}{black}

\newcommand{\mymacro}[1]{{\color{MacroColor} #1}}

\usepackage{mathtools}
\usepackage{ellipsis}
\makeatletter
\renewcommand*{\mathellipsis}{%
  \mathinner{%
    \kern\ellipsisbeforegap%
    {\ldotp}\kern\ellipsisgap%
    {\ldotp}\kern\ellipsisgap%
    {\ldotp}\kern\ellipsisaftergap%
  }%
}
\renewcommand*{\dotsb@}{%
  \mathinner{%
    \kern\ellipsisbeforegap%
    {\cdotp}\kern\ellipsisgap%
    {\cdotp}\kern\ellipsisgap%
    {\cdotp}\kern\ellipsisaftergap%
  }%
}
\renewcommand*{\@cdots}{%
  \mathinner{%
    \kern\ellipsisbeforegap%
    {\cdotp}\kern\ellipsisgap%
    {\cdotp}\kern\ellipsisgap%
    {\cdotp}\kern\ellipsisaftergap%
  }%
}
\renewcommand*{\ellipsis@default}{%
  \ellipsis@before
  \kern\ellipsisbeforegap
  .\kern\ellipsisgap
  .\kern\ellipsisgap
  .\kern\ellipsisgap
  \ellipsis@after\relax}
\renewcommand*{\ellipsis@centered}{%
  \ellipsis@before
  \kern\ellipsisbeforegap
  .\kern\ellipsisgap
  .\kern\ellipsisgap
  .\kern\ellipsisaftergap
  \ellipsis@after\relax}
\AtBeginDocument{%
  \DeclareRobustCommand*{\dots}{%
    \ifmmode\@xp\mdots@\else\@xp\textellipsis\fi}}
\def\ellipsisgap{-.05em}
\def\ellipsisbeforegap{-.05em}
\def\ellipsisaftergap{.05em}
\makeatother

\usepackage[normalem]{ulem} 
\usepackage{xcolor}         
\usepackage{bbm}

\newcommand{\defeq}{\mathrel{\overset{\raisebox{-0.25ex}{\textnormal{\tiny def}}}{=}}}

\newcommand{\indicator}[1]{\mathbbm{1}\mleft\{#1\mright\}}

\newcommand{\srcfont}[1]{{\color{SymbolColor}\texttt{#1}}}  

\newcommand{\unitAlphabet}[0]{{\color{UnitColor}U}}
\newcommand{\unitStrings}[0]{\kleene{\unitAlphabet}}

\newcommand{\srcAlphabet}{{\color{SymbolColor}\ensuremath{\Sigma}}\xspace}
\newcommand{\srcStrings}[0]{\kleene{\srcAlphabet}}
\newcommand{\tgtAlphabet}{{\color{SrcColor}\ensuremath{\Delta}}\xspace}
\newcommand{\tgtStrings}[0]{\kleene{\tgtAlphabet}}

\newcommand{\srcAlphabetOne}{\ensuremath{{\color{SymbolColor}\Sigma}_{1}}\xspace}
\newcommand{\srcAlphabetTwo}{\ensuremath{{\color{SymbolColor}\Sigma}_{2}}\xspace}
\newcommand{\srcAlphabetTwoStrings}{\ensuremath{{\color{SymbolColor}\Sigma}_{2}^{*}}\xspace}

\newcommand{\kleene}[1]{#1^*}
\newcommand{\kleeneplus}[1]{#1^+}

\newcommand{\bos}{{\mymacro{\textsc{bos}}}}
\newcommand{\eos}{{\mymacro{\textsc{eos}}}}

\newcommand{\myMintedFunc}[1]{\mbox{\texttt{{\color{blue}#1}}}}
\newcommand{\decomposeNext}[0]{\myMintedFunc{decompose\_next}}

\DeclareTextCommandDefault{\textvisiblespace}{%
  \mbox{\kern.06em\vrule \@height.3ex}%
  \vbox{\hrule \@width.3em}%
  \hbox{\vrule \@height.3ex}}

\renewcommand{\c}[0]{{\cdot}}

\definecolor{SymbolColor}{RGB}{140,10,89}  
\definecolor{WordColor}{RGB}{102,0,204}
\definecolor{UTFColor}{RGB}{102,178,255}

\definecolor{UnitColor}{RGB}{98,115,19}  
\definecolor{SrcColor}{RGB}{0,90,140}  
\colorlet{SepColor}{SrcColor}  

\newcommand{\tokenfont}[2][]{%
  \begin{tabular}{@{}c@{}}%
    {\color{SymbolColor}\texttt{#2}}\\[-5.5pt]%
    {\color{black!50}\scriptsize\texttt{#1}}%
  \end{tabular}%
}
\newcommand{\tgtfont}[1]{{\color{SrcColor}\strut\texttt{#1}}}  

\newcommand{\pChar}{\mymacro{p}}

\newcommand{\pSource}{\mymacro{p_{\srcAlphabet}}}

\newcommand{\pTarget}{\mymacro{p_{\tgtAlphabet}}}
\newcommand{\pUnit}{\pU}  
\newcommand{\pU}{\mymacro{p_{\unitAlphabet}}}

\newcommand{\pHuman}{{\mymacro{p_{\mathrm{H}}}}}

\newcommand{\prefixLanguageOp}[1]{\overrightarrow{#1}}
\newcommand{\prefixChar}{\mymacro{\prefixLanguageOp{\pChar}}}

\newcommand{\prefixTarget}{\mymacro{\prefixLanguageOp{\pTarget}}}

\let\prefixU\prefixUnit

\renewcommand{\Pr}[2][]{\mathop{\mathrm{Pr}}_{#1}\mleft[#2\mright]}

\newcommand{\tikzbracketuline}[1]{%
  \tikz[baseline=(t.base)]{
    \node[inner sep=0, outer sep=0, text depth=1.2pt] (t) {#1};
    \draw[line width=0.5pt] (t.south west) -- ++(0,-1pt);
    \draw[line width=0.5pt] ($(t.south west)+(0,-1pt)$) -- ($(t.south east)+(0,-1pt)$);
    \draw[line width=0.5pt] ($(t.south east)+(0,-1pt)$) -- (t.south east);
  }%
}
\newcommand{\underbracTarget}[1]{{\color{UnitColor}\tikzbracketuline{\textsc{#1}}}}

\newcommand{\huggingface}[0]{\emojiimg{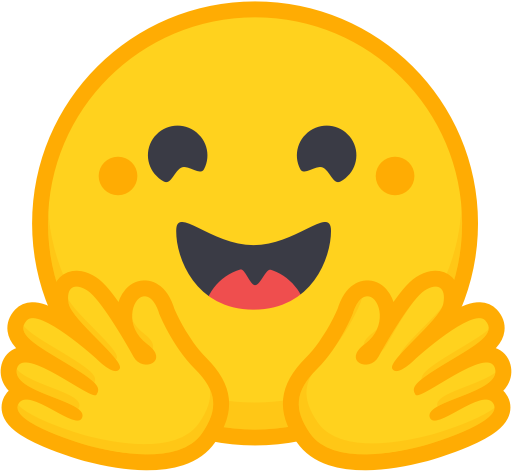}}

\newcommand{\utterance}[0]{\ensuremath{{\color{UnitColor}\boldsymbol{u}}}\xspace}
\newcommand{\parsingf}[0]{\ensuremath{{\color{UnitColor}\rho}}\xspace}  

\newcommand{\roi}{\utterance_{[i,j)}}

\newcommand{\SEP}{{\color{SepColor}\textsc{sep}}\xspace}

\newcommand{\sepAlphabet}{\ensuremath{{\color{SepColor}\Xi}}\xspace}
\newcommand{\sepStr}{\ensuremath{{\color{SepColor}\boldsymbol{\xi}}}\xspace}

\newcommand{\transFnSym}[0]{f}
\newcommand{\transFn}{\ensuremath{{\color{SrcColor}\transFnSym}}\xspace}

\newcommand{\transFnInv}{\ensuremath{{\color{SymbolColor}\transFnSym}^{-1}}\xspace}

\newcommand{\transHSym}[0]{h}
\newcommand{\transH}{\ensuremath{{\color{SrcColor}\transHSym}}\xspace}
\newcommand{\transHinv}{\ensuremath{{\color{UnitColor}\transHSym}^{-1}}\xspace}
\newcommand{\realization}{\ensuremath{{\color{SymbolColor}\rho}^{-1}}\xspace}

\newcommand{\unittext}[1]{{\color{UnitColor}\textsc{#1}}}
\newcommand{\symboltext}[1]{{\color{SymbolColor}\texttt{#1}}}
\newcommand{\tokid}[1]{{\color{black!50}\ttfamily #1}}

\newcommand{\transST}{{\color{SepColor}\texttt{f}}}

\newcommand{\STPTB}{{\transST}_{\mathrm{ptb}}}
\newcommand{\STLEAD}{{\transST}_{\mathrm{L}}}
\newcommand{\STTRAIL}{{\transST}_{\mathrm{T}}}

\newcommand{\psychometric}{\mymacro{r}}

\newcommand{\lltgt}{\text{LL}_{\target}}
\newcommand{\llb}{\text{LL}_{\baseline}}
\newcommand{\dll}{\Delta_{\text{llh}}}

\newcommand{\baseline}{\text{bl}}
\newcommand{\target}{\text{tgt}}

\newcommand{\delims}{{\color{SymbolColor}D}}

\newcommand{\unit}{{\color{UnitColor}{u}}}
\newcommand{\unitseq}{{\color{UnitColor}{\boldsymbol{u}}}}
\newcommand{\context}{\unitseq_{<t}}

\newcommand{\surprisal}{\mymacro{s}}

\newcommand{\srcSym}{{\color{SymbolColor}\sigma}}
\newcommand{\srcStr}
{\ensuremath{{\color{SymbolColor}\boldsymbol{\sigma}}}\xspace}

\newcommand{\tgtSym}{{\color{SrcColor}\delta}}
\newcommand{\tgtStr}{{\color{SrcColor}\boldsymbol{\delta}}}

\newcommand{\predvec}{\mymacro{\mathbf{x}}}

\newcommand{\States}{\mymacro{\mathcal{Q}}}
\newcommand{\Transitions}{\mymacro{\Omega}}
\newcommand{\FinalStates}{\mymacro{\mathcal{F}}}
\newcommand{\state}{\mymacro{q}}
\newcommand{\statep}{\state^{\prime}}

\newcommand{\InitialStates}[0]{\mymacro{\mathcal{I}}}
\newcommand{\wsp}{\mymacro{␣}}
\newcommand{\wspsym}{\mymacro{\text{\textvisiblespace}}}

\newcommand{\alphabet}{\mymacro{\srcAlphabet}}
\newcommand{\strings}{\mymacro{\alphabet^{*}}}

\newcommand{\significance}{p}

\usepackage{etoolbox}

\makeatletter
\newcommand{\appendixtableofcontents}{%
  \section*{Appendix Contents}%
  \@starttoc{atoc}%
}
\newcommand{\startappendixtoc}{%
  \let\oldaddcontentsline\addcontentsline
  \renewcommand{\addcontentsline}[3]{%
    \oldaddcontentsline{##1}{##2}{##3}
    \ifstrequal{##1}{toc}{\oldaddcontentsline{atoc}{##2}{##3}}{}
  }%
}
\makeatother

\usepackage{fancyvrb}
\crefname{FancyVerbLine}{line}{lines}
\crefname{figure}{Figure}{Figures}
\Crefname{figure}{Figure}{Figures}
\crefname{table}{Table}{Tables}
\Crefname{table}{Table}{Tables}
\crefname{proposition}{Proposition}{Propositions}
\Crefname{proposition}{Proposition}{Propositions}
\crefname{exx}{ex.}{ex.}
\Crefname{exx}{Ex.}{Ex.}
\crefname{ExNo}{ex.}{ex.}
\Crefname{ExNo}{Ex.}{Ex.}

\title{On the Proper Treatment of Units in Surprisal Theory}

\author{
  \textbf{Samuel Kiegeland}\textsuperscript{\srcAlphabet{\mdseries ,}\tgtAlphabet} \quad
  \textbf{Vésteinn Snæbjarnarson}\textsuperscript{\srcAlphabet{\mdseries ,}{\color{UnitColor}\ensuremath{U}}} \quad
  \textbf{Tim Vieira}\textsuperscript{\srcAlphabet} \quad
  \textbf{Ryan Cotterell}\textsuperscript{\srcAlphabet} \\
  \textsuperscript{\srcAlphabet}ETH Zürich \quad
  \textsuperscript{\tgtAlphabet}CHI-FRO \quad
  \textsuperscript{{\color{UnitColor}\ensuremath{U}}}University of Copenhagen \\
  \texttt{\{\href{mailto:samuel.kiegeland@gmail.com}{samuel.kiegeland}, \href{mailto:vest.snae@gmail.com}{vest.snae}, \href{mailto:tim.f.vieira@gmail.com}{tim.f.vieira}\}@gmail.com}\\
  \texttt{\href{mailto:ryan.cotterell@inf.ethz.ch}{ryan.cotterell@inf.ethz.ch}}}

\begin{document}
\maketitle
\begin{abstract}
Surprisal theory links human processing effort to the predictability of an upcoming linguistic unit, but empirical work often leaves the notion of a \emph{unit} underspecified. In practice, experimental stimuli are segmented into linguistically motivated units (e.g., words), while pretrained language models assign probability mass to a fixed token alphabet that typically does not align with those units. As a result, surprisal-based predictors depend implicitly on ad hoc procedures that conflate two distinct modeling choices: the definition of the unit of analysis and the choice of regions of interest over which predictions are evaluated. In this paper, we disentangle these choices and give a unified framework for reasoning about surprisal over arbitrary unit inventories. We argue that surprisal-based analyses should make these choices explicit and treat tokenization as an implementation detail rather than a scientific primitive.\looseness=-1

\vspace{.5em}
\hspace{1.25em}\includegraphics[width=1.25em,height=1.25em]{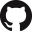}{\hspace{.75em}\parbox{\dimexpr\linewidth-2\fboxsep-2\fboxrule}{\url{https://github.com/samuki/units-surprisal}}}

\end{abstract}

\section{Introduction}
A long line of work in psycholinguistics has sought to characterize the processing difficulty that a comprehender experiences upon encountering a linguistic unit in context \citep[\textit{inter alia}]{miller-1964, ehrlich-et-al-1981-context, BALOTA1985364}.
A prominent \emph{computational} \citep{Marr-1982} account of such processing difficulty is surprisal theory \citep{hale-2001-probabilistic, levy2008expectation}, which posits that processing effort is determined by a unit's surprisal: the negative log-probability of encountering that unit given its preceding context.\footnote{This probability is understood as the comprehender's own predictive distribution over upcoming linguistic units, derived from an unobserved human language model. Empirical studies typically approximate this distribution with language models trained on natural language text.}

In early experimental work on surprisal theory, researchers often built and trained their own language models for a given dataset and experimental paradigm \citep[\textit{inter alia}]{hale-2001-probabilistic, levy2008expectation, demberg-2008, mitchell-etal-2010-syntactic, goodkind-bicknell-2018-predictive}.
Because they controlled the entire modeling pipeline, they were free to choose the basic units of the language model, i.e., its \emph{alphabet}.
For example, in his seminal work, \citet{hale-2001-probabilistic} trained a probabilistic context-free grammar over units derived from the Penn Treebank \citep{marcus-etal-1993-building}, i.e., units that follow the Penn Treebank's tokenization scheme.\footnote{In line with the fashion of the time, \citet{hale-2001-probabilistic} populated the model's vocabulary with high-frequency words from the training portion of the Penn Treebank, together with a distinguished out-of-vocabulary symbol.}
However, as language models grew---both in parameter count and in the size of their training corpora---it became inconvenient, if not infeasible, to train models from scratch for each study using the proper alphabet.

\begin{figure}[t]
\centering
\begin{tikzpicture}
\pgfmathsetmacro{\cw}{0.23}
\def\lx{0.5}
\def\spc{{\color{black!40}\textvisiblespace}}
\def\yT{0}        
\def\yC{-0.62}    
\def\ySep{-0.95}  
\def\yD{-1.28}    
\def\yP{-1.80}    
\def\ySepR{-2.13} 
\def\yR{-2.46}    
\tikzset{
    sbox/.style={rounded corners=2pt,
        inner xsep=0pt, inner ysep=0pt, anchor=west},
    tbox/.style={sbox, minimum height=5mm,
        draw=SymbolColor!50, fill=SymbolColor!6,
        font=\footnotesize},
    dbox/.style={sbox, minimum height=4.5mm,
        draw=UnitColor!50, fill=UnitColor!6,
        text height=1.5ex, text depth=0.35ex, font=\footnotesize},
    pbox/.style={sbox, minimum height=4.5mm,
        draw=UnitColor!50, fill=UnitColor!6,
        text height=1.5ex, text depth=0.35ex, font=\footnotesize},
    cbox/.style={sbox, minimum height=4.5mm,
        draw=SymbolColor!50, fill=SymbolColor!6,
        font=\fontsize{5.5}{7}\selectfont,
        text height=1.3ex, text depth=0.35ex},
    rlbl/.style={font=\sffamily\footnotesize, anchor=east},
    rbox/.style={sbox, minimum height=4.5mm,
        draw=UnitColor!50, fill=UnitColor!6,
        text height=1.5ex, text depth=0.35ex, font=\footnotesize},
}
\useasboundingbox ({\lx-0.90}, 0.48) rectangle ({\lx+25*\cw+0.48}, {\yR-0.28});

\node[rlbl, text=SymbolColor!70!black] at ({\lx-0.10}, \yT) {GPT-2};
\node[tbox, minimum width={6*\cw cm}] at ({\lx+0*\cw},  \yT) {%
    \begin{tabular}{@{}c@{}}{\srcfont{\strut Tokens}}\\[-5pt]{\fontsize{5}{6}\selectfont\tokid{\strut 22906}}\end{tabular}};
\node[tbox, minimum width={4*\cw cm}] at ({\lx+6*\cw},  \yT) {%
    \begin{tabular}{@{}c@{}}{\srcfont{\strut\spc don}}\\[-5pt]{\fontsize{5}{6}\selectfont\tokid{\strut 836}}\end{tabular}};
\node[tbox, minimum width={2*\cw cm}] at ({\lx+10*\cw}, \yT) {%
    \begin{tabular}{@{}c@{}}{\srcfont{\strut 't}}\\[-5pt]{\fontsize{5}{6}\selectfont\tokid{\strut 470}}\end{tabular}};
\node[tbox, minimum width={6*\cw cm}] at ({\lx+12*\cw}, \yT) {%
    \begin{tabular}{@{}c@{}}{\srcfont{\strut\spc equal}}\\[-5pt]{\fontsize{5}{6}\selectfont\tokid{\strut 4961}}\end{tabular}};
\node[tbox, minimum width={6*\cw cm}] at ({\lx+18*\cw}, \yT) {%
    \begin{tabular}{@{}c@{}}{\srcfont{\strut\spc words}}\\[-5pt]{\fontsize{5}{6}\selectfont\tokid{\strut 2456}}\end{tabular}};
\node[tbox, minimum width={1*\cw cm}] at ({\lx+24*\cw}, \yT) {%
    \begin{tabular}{@{}c@{}}{\srcfont{\strut .}}\\[-5pt]{\fontsize{5}{6}\selectfont\tokid{\strut 13}}\end{tabular}};

\node[rlbl, text=SymbolColor!70!black] at ({\lx-0.10}, \yC) {Chars};
\foreach \c/\i in {T/0,o/1,k/2,e/3,n/4,s/5,{\spc}/6,d/7,o/8,n/9,{'}/10,t/11,{\spc}/12,e/13,q/14,u/15,a/16,l/17,{\spc}/18,w/19,o/20,r/21,d/22,s/23,./24} {
    \node[cbox, minimum width={\cw cm}] at ({\lx+\i*\cw}, \yC) {\symboltext{\c}};
}

\draw[black!55, line width=0.5pt] ({\lx}, \ySep) -- ({\lx+25*\cw}, \ySep);

\node[rlbl, text=UnitColor!70!black] at ({\lx-0.10}, \yD) {Acont.};
\node[dbox, minimum width={6*\cw cm}] at ({\lx+0*\cw},  \yD) {\unittext{Tokens}};
\node[dbox, minimum width={5*\cw cm}] at ({\lx+7*\cw},  \yD) {\unittext{don't}};
\node[dbox, minimum width={5*\cw cm}] at ({\lx+13*\cw}, \yD) {\unittext{equal}};
\node[dbox, minimum width={6*\cw cm}] at ({\lx+19*\cw}, \yD) {\unittext{words.}};

\node[rlbl, text=UnitColor!70!black] at ({\lx-0.10}, \yP) {PTB};
\node[pbox, minimum width={6*\cw cm}] at ({\lx+0*\cw},  \yP) {\unittext{Tokens}};
\node[pbox, minimum width={2*\cw cm}] at ({\lx+7*\cw},  \yP) {\unittext{do}};
\node[pbox, minimum width={3*\cw cm}] at ({\lx+9*\cw},  \yP) {\unittext{n't}};
\node[pbox, minimum width={5*\cw cm}] at ({\lx+13*\cw}, \yP) {\unittext{equal}};
\node[pbox, minimum width={5*\cw cm}] at ({\lx+19*\cw}, \yP) {\unittext{words}};
\node[pbox, minimum width={1*\cw cm}] at ({\lx+24*\cw}, \yP) {\unittext{.}};

\draw[black!55, line width=0.5pt] ({\lx}, \ySepR) -- ({\lx+25*\cw}, \ySepR);

\node[rlbl, text=UnitColor!70!black] at ({\lx-0.10}, \yR) {ROIs};
\node[rbox, minimum width={6*\cw cm}] at ({\lx+0*\cw}, \yR) {};
\node[anchor=center, font=\footnotesize, text=UnitColor!60!black]
    at ({\lx+3*\cw}, \yR) {\unittext{NP}};
\node[rbox, minimum width={17*\cw cm}] at ({\lx+7*\cw}, \yR) {};
\node[anchor=west, font=\footnotesize, text=UnitColor!60!black]
    at ({\lx+7.2*\cw}, \yR) {\unittext{VP}};
\node[sbox, minimum height=3.2mm, minimum width={11*\cw cm},
      draw=UnitColor!50, densely dashed, fill=UnitColor!12,
      rounded corners=1.5pt]
      at ({\lx+13*\cw}, \yR) {};
\node[anchor=west, font=\fontsize{6}{7}\selectfont, text=UnitColor!55!black]
    at ({\lx+13.2*\cw}, \yR) {\unittext{VP}};
\node[rbox, minimum width={1*\cw cm}] at ({\lx+24*\cw}, \yR) {};
\node[anchor=center, font=\footnotesize, text=UnitColor!60!black]
    at ({\lx+24.5*\cw}, \yR) {\unittext{.}};

\node[anchor=west, font=\footnotesize] at ({\lx+25*\cw+0.12}, {(\yT+\yC)/2}) {$\srcAlphabet$};
\node[anchor=west, font=\footnotesize] at ({\lx+25*\cw+0.12}, {(\yD+\yP)/2}) {$\unitAlphabet$};
\end{tikzpicture}
\caption{The string \textit{Tokens don't equal words.}\ at three levels: two alphabets of symbols $\srcAlphabet$, two unit inventories $\unitAlphabet$, and regions of interest (ROIs) derived from the sentence's constituency parse: NP, VP (with a nested inner VP shown dashed), and punctuation.
The contraction \textit{don't} is split three ways: \textcolor{SymbolColor}{GPT-2} yields \srcfont{don}\,$|$\,\srcfont{'t}, \textcolor{UnitColor}{Penn Treebank (PTB)} yields \unittext{do}\,$|$\,\unittext{n't}, and the \textcolor{UnitColor}{acontextual} inventory keeps it as one unit \unittext{don't}. The period in \unittext{words.} is similarly attached in the acontextual inventory but separate in the contextual one.
}
\label{fig:token-unit-mismatch}
\end{figure}
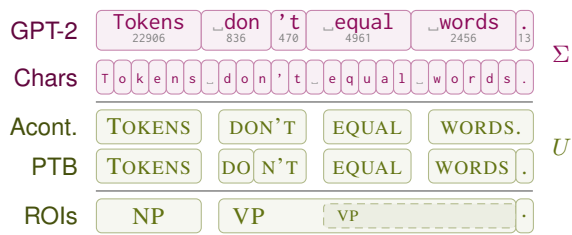

With the shift toward large pretrained models \citep{wilcox2020predictive, wilcox-etal-2023-testing, oh-schuler-2023-surprisal}, experimenters are no longer free to attune the model’s alphabet to their datasets. Instead, they inherit a fixed tokenization, e.g., byte-pair encoding \citep{gage-1994-a, sennrich-etal-2016-neural}, whose units generally do not coincide with linguistically meaningful ones \citep{Church_2020, hofmann-etal-2021-superbizarre, nair-resnik-2023-words}.
This mismatch has given rise to a methodological infelicity. Researchers must reconcile the gap between their desired units and the model’s alphabet (see \cref{fig:token-unit-mismatch}), typically through bespoke post hoc procedures that impose unit boundaries on token strings \citep{wilcox2020predictive, nair-resnik-2023-words, wilcox-etal-2023-testing, pimentel-meister-2024-compute, oh-schuler-2024-leading}. Such heuristics vary widely between papers and muddle the interpretation of the units.

Units are not the only level at which we may seek an analysis.
For instance, one may wish to relate surprisal to discourse structure \citep{tsipidi-etal-2024-surprise, tsipidi-etal-2025-harmonic} by aggregating word-level surprisals into a discourse-level region of interest \citep[ROI;][]{giulianelli-etal-2024-proper}.
Beyond the choice of units, then, the modeler must also determine the ROIs, i.e., how units are aggregated into predictors.
Units and ROIs often coincide one-to-one, but need not: ROIs may span multiple units or even overlap, unlike units themselves.
One might \textit{a priori} construct a language model with discourse-level units, but this requires a more permissive definition than the one given here, because discourse structures are nested like syntactic constituents.
See \cref{fig:token-unit-mismatch} for an example of nested constituent ROIs.

This paper seeks to promote an understanding of the proper treatment of units and ROIs in surprisal theory, and provides a practical toolkit that enables modelers to select their desired units freely.
Our proposal is straightforward: the experimenter first chooses the unit inventory best suited for their analysis.
Then, if the language model cannot be retrained, it should be converted to the chosen alphabet, e.g., by composing the language model with an appropriate function \citep{snbjarnarson2026transducing}.
On this view, tokenization is little more than an implementation detail that should be of no scientific importance; the unit of analysis is a modeling choice and should be selected to match the scientific question at hand.
Depending on the goal, units may be taken to be the model’s \emph{token} alphabet \citep{beinborn-pinter-2023-analyzing, nair-resnik-2023-words}, they can be defined by an explicit segmentation scheme, such as simple delimiter rules \citep{oh-schuler-2024-leading,pimentel-meister-2024-compute}, or even derived using contextual segmentation rules, such as the Penn Treebank guidelines \citep[PTB;][]{marcus-etal-1993-building} or Universal Dependencies \citep[UD;][]{nivre-etal-2017-universal}.\looseness=-1

\section{Language Models}
Let $\srcAlphabet$ be an \defn{alphabet}, i.e., a finite, non-empty set whose elements are called \defn{symbols}.
A \defn{string} over $\srcAlphabet$ is a finite sequence of symbols $\srcStr = \srcSym_1 \srcSym_2 \cdots \srcSym_T$ with $\srcSym_t \in \srcAlphabet$.
We write $\strings$ for the set of all strings over $\srcAlphabet$, including the \defn{empty string} $\varepsilon$, and $\kleeneplus{\srcAlphabet} \defeq \srcStrings \setminus \{\varepsilon\}$.
Furthermore, we write $\eos\notin\srcAlphabet$ for a distinguished \defn{end-of-sequence} symbol.
We use $\srcStr \c \srcStr^{\prime}$ to denote the \defn{concatenation} of $\srcStr, \srcStr^{\prime} \in \strings$ and write $\srcStr \preceq \srcStr^{\prime}$ if $\srcStr$ is a prefix of $\srcStr^{\prime}$.

A \defn{language model} $\pSource$ is a probability distribution over $\srcStrings$.  
For any $\srcStr \in \srcStrings$, $\pSource(\srcStr)$ factors as
\begin{equation}
\pSource(\srcStr) = \prefixChar(\eos \mid \srcStr) \prod_{t = 1}^{T} \prefixChar\mleft(\srcSym_t \mid \srcStr_{<t}\mright),
\end{equation}
where the \defn{conditional prefix probability} is
\begin{align}
\label{eq:ratio}
\prefixChar(\srcStr' \mid \srcStr) 
&\defeq
\Pr[Y \sim \pSource]{ Y \succeq \srcStr \c \srcStr^{\prime} \,\middle|\, Y \succeq \srcStr }
\\ &= \frac{\prefixChar(\srcStr\c \srcStr^{\prime})}{\prefixChar(\srcStr)},
\end{align}
and the \defn{prefix probability} is 
\begin{align}
\label{eq:prefix-prob}
\prefixChar(\srcStr) &\defeq 
\Pr[Y \sim \pSource]{ Y \succeq \srcStr} 
\\ &= \sum_{ \srcStr' \in \srcStrings } \indicator{\srcStr^{\prime} \succeq \srcStr} \,  \pSource(\srcStr^{\prime}).
\end{align}

\section{Units}
\label{sec:units}
From the cognitive perspective, linguistic utterances are generally taken to be divisible into discrete units.\footnote{The claim that speech is divisible into discrete units is itself an idealization. The continuous acoustic signal is carved into discrete segments by convention; the mapping from articulatory gestures to perceived phonemes involves substantial coarticulation and context-dependence \citep{liberman1967perception}.\looseness=-1}
The question of what constitutes a linguistic unit has been debated at least since \citet{Saussure1997clg2} and \citet{bloomfield1933language}, and remains contentious to this day \citep{haspelmath2011indeterminacy, murphy2024word}.
Informally, a unit is a discrete segment of linguistic structure---a phoneme, morpheme, word, or phrase---that serves as an atom of analysis at a chosen level of description.
It is generally agreed that linguistic units exist at various levels, i.e., utterances can be divided into clauses, clauses into words, words into morphemes, morphemes into phonemes, and phonemes into phones---even while the notion of a word is a hotly contested one \citep{haspelmath2011indeterminacy, dixon2002word}.
Each of these levels provides a valid granularity at which one can decompose an utterance \citep{hockett1958course}.
The choice of level is not merely a technical nuisance but a substantive commitment about the granularity at which cognitive processing is modeled.
When---as is common in practice---the model's alphabet is taken to coincide with the unit inventory, this choice directly determines the events to which
probability is assigned and, consequently, the quantities that enter the linking hypothesis relating surprisal to behavioral data.

Despite this importance, the choice of units has received comparatively little attention in surprisal theory.
In practice, most studies inherit their units from the tokenizer of a pretrained language model or from the segmentation conventions of a particular eye-tracking corpus, treating the unit inventory as fixed rather than a variable to be controlled. This obscures a logically prior question: what \emph{should} the units be?
Answering it requires separating $\unitAlphabet$, the countable (but not necessarily finite) unit inventory over which we wish to define surprisal, from $\srcAlphabet$, the alphabet of the language model.
We make this separation explicit below. Rather than advocating for a specific unit inventory, we develop a general formalism---in a slogan, ``bring your own units''---that is compatible with any choice the modeler wishes to make, irrespective of the language model's native alphabet.
To that end, we assume the modeler has a countable unit inventory $\unitAlphabet$.
This set may be finite, e.g., the set of phonemes in a language, or countably infinite, e.g., the set of orthographic words.\footnote{In addition to neologism, recursive morphological processes, e.g., compounding and derivation, produce unboundedly many distinct word forms, yielding a countably infinite set of orthographic words; see \citet[Ch.~5]{pinker1994language}.}

A linguistic \defn{utterance} is a string of units $\utterance \in \unitStrings$.
A \defn{unit parser} is a stochastic map $\parsingf \colon \srcStrings \rightsquigarrow \unitStrings$ that maps each symbol string $\srcStr \in \srcStrings$ to a probability distribution $\parsingf(\cdot \mid \srcStr)$ over unit strings. In many languages, parsing into units is inherently ambiguous; we give a canonical example of such ambiguity from Mandarin Chinese.
\begin{myexample}\label{ex:ambiguity}
Let $\srcAlphabet$ be the set of Chinese characters, and let $\unitAlphabet = \kleeneplus{\srcAlphabet}$.
Consider the string $\srcStr = {}${\begin{CJK}{UTF8}{gbsn}\symboltext{乒乓球拍卖完了}\end{CJK}}. The unit parser $\parsingf$ ought to assign positive probability to (at least) two unit strings:

\ex.\label{ex:chinese-parses}
    \a. $\parsingf(\srcStr) = {}${\color{UnitColor}\begin{CJK}{UTF8}{gbsn}乒乓球\end{CJK}}\,$\cdot$\,{\color{UnitColor}\begin{CJK}{UTF8}{gbsn}拍卖\end{CJK}}\,$\cdot$\,{\color{UnitColor}\begin{CJK}{UTF8}{gbsn}完了\end{CJK}}\\[2pt]
    {\small\textit{p\={\i}ngp\=angqi\'u}\quad\textit{p\=aim\`ai}\quad\textit{w\'anle}}\\[1pt]
    {\small ping pong ball\quad auction\quad finish-\textsc{perf}}\\[1pt]
    {\small ``The ping-pong ball auction is over.''}
    \b. $\parsingf(\srcStr) = {}${\color{UnitColor}\begin{CJK}{UTF8}{gbsn}乒乓球拍\end{CJK}}\,$\cdot$\,{\color{UnitColor}\begin{CJK}{UTF8}{gbsn}卖\end{CJK}}\,$\cdot$\,{\color{UnitColor}\begin{CJK}{UTF8}{gbsn}完了\end{CJK}}\\[2pt]
    {\small\textit{p\={\i}ngp\=angqi\'up\=ai}\quad\textit{m\`ai}\quad\textit{w\'anle}}\\[1pt]
    {\small ping-pong-paddle\quad sell\quad finish-\textsc{perf}}\\[1pt]
    {\small ``The ping-pong paddles have been sold.''}

\noindent The character {\begin{CJK}{UTF8}{gbsn}\symboltext{拍}\end{CJK}} (\textit{p\=ai}) is the pivot: it can end the compound {\color{UnitColor}\begin{CJK}{UTF8}{gbsn}球拍\end{CJK}} (\textit{qi\'up\=ai}, `paddle') or begin {\color{UnitColor}\begin{CJK}{UTF8}{gbsn}拍卖\end{CJK}} (\textit{p\=aim\`ai}, `auction'), yielding different unit strings from the same symbol string.
\end{myexample}

\noindent 
Our choice of a Chinese example is strategic.
In many languages that use whitespace in their orthography, e.g., English and most other European languages, parsing into units is far more deterministic---for example, the Penn Treebank tokenization convention \citep{marcus-etal-1993-building} assigns a unique segmentation to every string.
In what follows, we therefore make the simplifying assumption that $\parsingf$ is \emph{deterministic}: for every $\srcStr \in \srcStrings$, there is exactly one unit string $\utterance$ with $\parsingf(\utterance \mid \srcStr) = 1$. Under this assumption, $\parsingf$ reduces to a total function $\srcStrings \to \unitStrings$.
In the remainder of the paper, we write $\parsingf(\srcStr) = \utterance$ and treat $\parsingf$ as a function. 
We call the unit parser's inverse its \defn{realization} $\realization \subseteq \unitStrings \times \srcStrings$.
In general, the realization is a relation, because one unit string may correspond to multiple symbol strings; this is true even in English.\looseness=-1

\begin{myexample}\label{ex:unit_system}
Let $\srcAlphabet$ be the set of cased ASCII characters, and let $\unitAlphabet = \{\unittext{Hale}, \unittext{cited}, \unittext{Levy}\}$. 
Assuming $\parsingf$ is deterministic, an English unit parser ought to map the following two symbol strings

\ex.\label{ex:multiple-realizations}
    \a. $\symboltext{Hale\wsp cited\wsp Levy.}$
    \b. $\symboltext{Hale\wsp cited\wsp\wsp\wsp Levy.}$
    
\noindent to the same unit string $\unittext{Hale} \c \unittext{cited} \c \unittext{Levy}$ as they only differ in terms of whitespace.
Thus, the realization $\realization$ is a non-functional relation.
\end{myexample}

\subsection{Pushforwards}
\label{sec:tlm}
Suppose we have a language model $\pSource$ over the alphabet $\srcAlphabet$, and we wish to define a language model $\pU$ over the unit inventory $\unitAlphabet$ via the unit parser $\parsingf \colon \srcStrings \rightarrow \unitStrings$.
The \defn{pushforward} of $\pSource$ through $\parsingf$ is given by the following expression
\begin{equation}\label{eq:pushforward}
\pU(\utterance) \defeq \sum_{\srcStr \in \realization(\utterance)} \pSource(\srcStr),
\end{equation}
where the sum ranges over all symbol strings $\srcStr$ that the unit parser maps to $\utterance$.
For $\pU$ to be a well-defined probability distribution, we require that $\parsingf$ be a total function $\srcStrings \to \unitStrings$, i.e., every symbol string $\srcStr \in \srcStrings$ is mapped to exactly one unit string.
Equivalently, the fibers
$\{ \{\srcStr \mid (\srcStr, \utterance) \in \parsingf\} \mid \utterance \in \unitStrings \}$ partition $\srcStrings$, which is always true for the fibers of a total function.
Without further assumptions on $\parsingf$, \cref{eq:pushforward} is difficult to compute---the number of symbol strings related to $\utterance$ may be countably infinite.\looseness=-1

\subsection{Lost in Whitespace}\label{sec:lost-in-whitespace}
Recent work \citep{oh-schuler-2024-leading,pimentel-meister-2024-compute} proposes a formalism for computing the probability of the next unit in context given a symbol-level language model.
We identify two conceptual issues with their shared approach---one mathematical and one linguistic.

\paragraph{Unit Inconsistency.}
Both papers formalize the conversion from a symbol-level language model to a unit-level model under three assumptions about the realization $\realization \subseteq \unitStrings \times \srcStrings$:
\begin{enumerate*}[label=(\roman*)]
\item $\realization$ is a monoid homomorphism,\footnote{The monoids in question are $(\unitStrings, \c, \varepsilon)$ and $(\srcStrings, \c, \varepsilon)$, i.e., the free monoids over $\unitAlphabet$ and $\srcAlphabet$ respectively, with concatenation as the binary operation and the empty string as the identity.} i.e., $\realization$ is a function where $\realization(\unit_1 \c \cdots \c \unit_T) = \realization(\unit_1) \c \cdots \c \realization(\unit_T)$; that is, each unit is realized in $\srcAlphabet$ independently, and the realization of a sequence of units is the concatenation of the individual realizations;
\item $\srcAlphabet$ can be partitioned into two disjoint subsets $\srcAlphabetOne$ and $\srcAlphabetTwo$, where $\srcAlphabetOne$ contains symbols that mark a unit boundary and $\srcAlphabetTwo$ does not;
\item each unit maps to a symbol string in $\srcAlphabetOne \circ \srcAlphabetTwoStrings$, i.e., one boundary-marking symbol followed by zero or more continuation symbols.
\end{enumerate*}
Assumption~(i) implies that $\realization$ is a function.
Assumption~(ii) is motivated by the fact that many tokenizers prepend a whitespace character to the first token of each word, so that \textvisiblespace{} signals a unit boundary.\footnote{\label{fn:bos}A notable exception occurs at the beginning of a string: the first symbol belongs to $\srcAlphabetTwo$ regardless of the unit it represents.
One can encode this by introducing an additional copy of all symbols in $\srcAlphabetOne$ with a distinguished $\bos$ symbol 
prepended.\looseness=-1}
Consider the two symbol strings

\ex.\label{ex:unit_inconsistency}
    \a. \symboltext{Hale\wsp cited\wsp Levy.}
    \b. \symboltext{Levy\wsp cited\wsp Hale.}

\noindent which we would naturally expect to correspond to the following unit strings

\ex.\label{ex:expected_units}
    \a. \unittext{Hale}$_1$ $\cdot$ \unittext{cited} $\cdot$ \unittext{Levy}
    \b. \unittext{Levy} $\cdot$ \unittext{cited} $\cdot$ \unittext{Hale}$_2$

where \unittext{Hale}$_1$ = \unittext{Hale}$_2$.
However, assumptions (i)--(iii) make such an equivalence impossible.
In the first string, \unittext{Hale}$_1$ is string-initial so $\realization(\unittext{Hale}_1) = \text{\bos\symboltext{Hale}}$, where $\bos$ is the beginning-of-string symbol from Footnote~\ref{fn:bos}.
In the second, \unittext{Hale}$_2$ is preceded by $\symboltext{\wsp}$ so $\realization(\unittext{Hale}_2) = \text{\wsp\symboltext{Hale}}$.
Because assumptions (i)--(iii) force $\realization$ to be a function, we have $\unittext{Hale}_1 \neq \unittext{Hale}_2$.
Consequently, \unittext{Hale} is forced to be two distinct units depending on its context.
Thus, the formalism cannot provide a coherent language model over units---the identity of a unit should not depend on whether it begins a string.
Our framework sidesteps this problem because $\realization$ is a relation, not a function: the unit \unittext{Hale} can stand in the realization relation to \emph{both} $\text{\bos\symboltext{Hale}}$ and $\text{\wsp\symboltext{Hale}}$, so a single unit suffices regardless of context.\looseness=-1

\paragraph{Linguistic Adequacy.}
A second concern is linguistic.
The approach of \citet{oh-schuler-2024-leading} and \citet{pimentel-meister-2024-compute} implicitly assumes that units can be recovered by grouping symbols in $\srcAlphabet$ according to the partition $\srcAlphabetOne \sqcup \srcAlphabetTwo$ from assumption~(ii).
But byte-pair encoding is a compression algorithm: the boundaries it induces are artifacts of corpus frequency, not of morphological or syntactic structure.
Moreover, their framework requires every symbol to be classified as either a boundary marker (in $\srcAlphabetOne$) or a continuation symbol (in $\srcAlphabetTwo$), uniformly across all contexts.
Yet a comma is word-internal in \symboltext{1,000} but marks a clause boundary in \symboltext{end, he said}; an apostrophe is word-internal in \symboltext{don't} but possessive-marking in \symboltext{cat's}. No fixed partition can capture such distinctions in general, since the same character serves different roles in different environments.
We remark that this concern is far from pedantic; \citet[\S2.2.1]{clark-2025} report discarding stimuli as the method does not properly handle punctuation. 

\subsection{Regular Unit Inventories}\label{sec:regularity}
A technical challenge arises when the unit inventory is infinite, because language models operate over finite alphabets by definition, and
the unit inventory $\unitAlphabet$ may be countably infinite, e.g., the set of all whitespace-delimited words.
The key observation is that even an infinite $\unitAlphabet$ can be finitely represented whenever each unit is itself a string over some finite alphabet, i.e., $\unitAlphabet \subseteq \kleene{\sepAlphabet}$, and $\unitAlphabet$ forms a \emph{regular} subset of $\kleene{\sepAlphabet}$.\footnote{A language is \emph{regular} if it is accepted by a finite automaton; see \citet{pin2010mathematical} for a comprehensive treatment. Also, note that regularity is a \emph{sufficient}, but not necessary condition.}
We call this the \defn{regularity assumption}. 

For many abstract linguistic units, the regularity assumption is well-motivated, e.g., it is widely established that phonotactic constraints are regular \citep{kaplan-kay-1994-regular, heinz-2018-computational}, as are many morphotactic rules \citep{koskenniemi-1983-two-level, beesley-karttunen-2003-finite}.\footnote{A notable exception is reduplication, which is not regular and thus falls outside the scope of standard finite-state methods. However, \citet{dolatian-heinz-2018-modeling} argue that 2-way finite transducers can nonetheless model reduplication.}
Because units at any of these levels are defined by regular constraints, regularity of $\unitAlphabet$ is a mild assumption in practice.
Under this assumption, we can reduce operations on a potentially infinite $\unitAlphabet$ to operations over regular sets as follows.
Let $\sepStr\in \kleene{\sepAlphabet}$, and let $\SEP \not\in \sepAlphabet$ be a distinguished separator symbol.
We define
\begin{equation}\label{eq:unit-homomorphism}
\begin{aligned}
\transH \colon &\unitAlphabet \to \kleene{\sepAlphabet}\SEP\\
&\unit \mapsto \sepStr\SEP,
\end{aligned}
\end{equation}which appends $\SEP$ to each unit's underlying string.\footnote{It will be important that $\transH$ be \defn{prefix free}, i.e., there do not exist units $\unit_1, \unit_2 \in \unitAlphabet$ such that $\transH(\unit_1) \preceq \transH(\unit_2)$. \label{fn:prefix-free}}
This extends to a monoid homomorphism $\unitStrings \rightarrow \kleene{(\kleene{\sepAlphabet}\SEP)}$ by defining $\transH(\unit_1 \cdot \cdots \cdot \unit_T) \defeq \transH(\unit_1) \cdot \cdots \cdot \transH(\unit_T)$.
Note that $\transH(\unitAlphabet)$
is regular since $\unitAlphabet$ is regular by assumption and regular sets are closed under concatenation, and so $\transH(\unitStrings) = \kleene{\transH(\unitAlphabet)}$ is regular by standard closure properties.

\subsection{Transduced Language Models}\label{sec:transduced-lms}

We now introduce a \emph{computational} formalism for describing $\parsingf \colon \srcStrings \rightarrow \unitStrings$.
First, let $\tgtAlphabet \defeq \sepAlphabet \sqcup \{\SEP\}$, following \Cref{eq:unit-homomorphism}.
Then, note that the composition $\transH \circ \parsingf \colon \srcStrings \to \kleene{(\sepAlphabet \sqcup \{\SEP\})}$ takes $\srcAlphabet$-strings and maps them to strings over the finite alphabet $\sepAlphabet \sqcup \{\SEP\}$.
If we can compute $\transH \circ \parsingf$, we can apply $\transHinv$ to the output to map back to a unit string.

A \defn{transducer} is a state machine encoding a string-to-string relation $\transFn \subseteq \srcStrings \times \tgtStrings$.
Formally, it is defined as a tuple
$\transST=(\States,\srcAlphabet, \tgtAlphabet, \Transitions, \InitialStates, \FinalStates)$, where $\States$ is a set of states,\footnote{In general, $\States$ need not be finite and may encode auxiliary memory (e.g., a stack), modeling context-free behavior.} $\srcAlphabet$ and $\tgtAlphabet$ are the input and output alphabets,
$\InitialStates,\FinalStates\subseteq\States$ are the sets of initial and final states, and
$\Transitions\subseteq \States\times(\srcAlphabet\cup\{\varepsilon\})\times(\tgtAlphabet \cup\{\varepsilon\})\times\States$ is the set of transitions.
For any two states $\state, \statep \in \States$, we write $(\state,\srcSym,\tgtSym,\statep)\in\Transitions$ as shorthand for the transition $\state \xrightarrow{\srcSym:\tgtSym} \statep$.
A transducer is called \defn{finite} when $\States$ is a finite set.\footnote{Finite transducers are widely used for representing string-to-string functions in natural language processing \citep{roche97finit_state, mohri-1997-finite}, and their mathematical properties are well-understood; see \citet{pin2010mathematical, Pin2021Handbook} for standard operations such as composition and determinization.
Here, we focus on rational functions, as they provide a compact framework for manipulating language models.\looseness=-1
}
A function is called \defn{rational} if it can be realized by a finite transducer.

Define $\transFn \defeq \transH \circ \parsingf \colon \srcStrings \to \tgtStrings$.
Note that $\transFn$ maps between two \emph{finite} alphabets, $\srcAlphabet$ and $\tgtAlphabet$, even when the unit inventory $\unitAlphabet$ is infinite\footnote{There is a relevant line of work that extends classical finite automata and transducers with transitions labeled by predicates over potentially infinite alphabets, known as \emph{symbolic} finite automata and transducers \citep{vannoord2003predicates, veanes2012symbolic, dantoni2017power}.} because $\unitAlphabet \subseteq \kleene{\sepAlphabet}$.
If, in addition, $\transFn$ is rational,\footnote{The regularity assumption is necessary for rationality of $\transFn$ whenever $\parsingf$ surjects onto $\unitStrings$.} \citet{snbjarnarson2026transducing} provide a practical algorithm for computing the pushforward (see \cref{sec:tlm}) under $\transFn$, defined as
\begin{equation}\label{eq:pushforward-decomp}
\pTarget(\tgtStr) \defeq \sum_{ \srcStr \in \transFnInv(\tgtStr)} \pSource(\srcStr).
\end{equation}
This gives us a distribution over $\tgtAlphabet$-strings; we now extract unit-level probabilities from it.
Writing $\prefixU$ for the prefix probability of $\pU$, the conditional next-unit probability is given by
\begin{equation}\label{eq:next-unit}
\prefixU(\unit \mid \utterance) = \prefixTarget(\transH(\unit) \mid \transH(\utterance)).
\end{equation}
The simplicity of \Cref{eq:next-unit} follows from the injectivity of $\transH$, by construction, and the fact that $\transH$ is prefix-free; see Footnote~\ref{fn:prefix-free}, and \cref{app:trailing-h} for discussion.

\begin{figure}[t]
\centering
\begin{tikzpicture}[
    >={Stealth[length=1.8mm,width=1.5mm]},
    every node/.style={font=\footnotesize},
    space/.style={align=center, inner sep=1pt},
    maplbl/.style={font=\footnotesize, inner sep=2pt},
]
\def\Sy{1.4}     
\def\By{-0.45}   
\def\Bx{2.8}     
\def\sep{0.10}   
\def\pull{0.22}  
\pgfmathsetmacro{\DL}{sqrt(\Bx*\Bx + (\Sy-\By)*(\Sy-\By))}
\pgfmathsetmacro{\ux}{\Bx/\DL}
\pgfmathsetmacro{\uy}{(\Sy-\By)/\DL}
\pgfmathsetmacro{\spx}{-\sep*\uy}    
\pgfmathsetmacro{\spy}{\sep*\ux}     
\pgfmathsetmacro{\pux}{\pull*\ux}    
\pgfmathsetmacro{\puy}{\pull*\uy}    

\node[space] (S) at (0, \Sy) {$\kleene{\tgtAlphabet}$};
\node[space] (X) at (-\Bx, \By) {$\srcStrings$};
\node[space] (V) at (\Bx, \By) {$\kleene{\unitAlphabet}$};

\draw[->] ($(X.north east)+(\spx,\spy)+(\pux,\puy)$) --
          ($(S.south west)+(\spx,\spy)$)
  node[maplbl, midway, above, xshift=-3pt, yshift=0pt] {$\transFn$};
\draw[->] ($(S.south west)+(-\spx,-\spy)+(-\pux,-\puy)$) --
          ($(X.north east)+(-\spx,-\spy)$)
  node[maplbl, midway, below, xshift=15pt, yshift=8pt] {$\transFnInv$};

\draw[->] ($(S.south east)+(\spy,-\spx)+(\pux,-\puy)$) --
          ($(V.north west)+(\spy,-\spx)$)
  node[maplbl, midway, above, xshift=3pt, yshift=4pt] {$\transHinv$};
\draw[->] ($(V.north west)+(-\spy,\spx)+(-\pux,\puy)$) --
          ($(S.south east)+(-\spy,\spx)$)
  node[maplbl, midway, below, xshift=-5pt] {$\transH$};

\draw[->, densely dotted] ($(X.east)+(1.7*\pull,\sep)$) -- ($(V.west)+(0,\sep)$)
  node[maplbl, midway, above, xshift=-3pt] {$\parsingf$};
\draw[->, densely dotted] ($(V.west)+(-\pull,-\sep)$) -- ($(X.east)+(0,-\sep)$)
  node[maplbl, midway, below] {$\realization$};
\end{tikzpicture}
\caption{The unit parser $\parsingf \colon \srcStrings \to \kleene{\unitAlphabet}$ passes through $\kleene{\tgtAlphabet}$: the transducer $\transFn$ maps symbol strings to $\SEP$-annotated strings, and $\transHinv$ splits on $\SEP$ and maps each segment to the unit it spells, recovering the unit string.
\vspace{-14pt}
}
\label{fig:realization-diagram}
\end{figure}
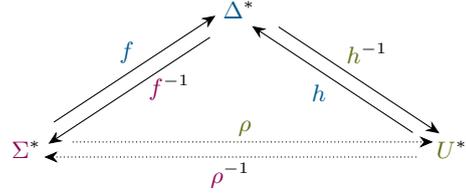

\section{Regions of Interest}\label{sec:roi}
Experimenters often study predictions at spans that cover multiple units, such as sentences \citep{lau-2017-sentences, meister-etal-2021-revisiting, giulianelli2023information}, dialogue turns \citep{wallbridge22_interspeech, wallbridge-etal-2023-dialogue}, or discourse segments \citep{tsipidi-etal-2024-surprise, tsipidi-etal-2025-harmonic}, each of which spans multiple word-level units.
To describe such spans formally, let $\kleeneplus{\unitAlphabet}=\unitStrings\setminus\{\varepsilon\}$. Given an utterance $\utterance=\unit_1\cdots\unit_T\in \kleeneplus{\unitAlphabet}$ of $T$ units, we write $\utterance_{[i,j)}=\unit_i\cdots\unit_{j-1}$ for $1\leq i < j \leq T+1$ and refer to it as a \defn{region of interest}\label{def:roi} \citep[ROI;][]{giulianelli-etal-2024-proper}.\footnote{Also referred to as an area of interest. \citep{schotter2025area}.}\looseness=-1

To predict reading time for an ROI $\roi$ from a language model $\pU$ over $\utterance \in \unitStrings$, one must specify how to combine unit-level surprisals.
The standard approach is to sum surprisals over the ROI \citep{smith2013, nair-resnik-2023-words}. Importantly, this sum yields the surprisal of the character sequence but omits the probability of $\SEP$ or $\eos$ signaling the unit boundary, so it is not directly comparable with unit-level surprisal.

Computing ROI-level surprisal requires that their boundaries be compatible with finer-grained unit boundaries; otherwise, a unit may overlap two ROIs and cannot be unambiguously assigned to either. For example, consider the stimulus \textit{Predictive power} with characters as units. A psycholinguist studying parafoveal preview \citep{rayner1975parafoveal, rayner1982availability, blanchard-etal-1989-acquisition} might define each ROI as the first three characters of a word. However, as shown in \Cref{ex:parafoveal_mismatch}, the ROI \underbracTarget{P$\cdot$r$\cdot$e} spans parts of two GPT-2 \citep{radford2019language} tokens (\srcfont{P} and \srcfont{redict}), so the model's token-level probabilities alone cannot yield the surprisal of this character span. To resolve such mismatches, we can transform a language model from its native token alphabet to the character alphabet \citep{pmlr-v267-vieira25a}.\looseness=-1

\ex.\label{ex:parafoveal_mismatch}
    \a. \underbracTarget{P$\cdot$r$\cdot$e}\unittext{$\cdot$d$\cdot$i$\cdot$c$\cdot$t$\cdot$i$\cdot$v$\cdot$e$\cdot$\wsp{}$\cdot$}\underbracTarget{p$\cdot$o$\cdot$w}\unittext{$\cdot$e$\cdot$r} \\ {\footnotesize Character units (ROIs underlined)}
    \b. \tokenfont[47]{P}\;\tokenfont[17407]{redict}\;\tokenfont[425]{ive}\;\tokenfont[1176]{\wsp{}power} \hfill {\footnotesize GPT-2 tokens}

\section{Surprisal Theory}
Surprisal theory \citep{hale-2001-probabilistic, levy2008expectation} posits that the incremental processing difficulty of language comprehension is a function of how unexpected an upcoming linguistic unit is given its context. The theory assumes an implicit human language model $\pHuman$, and predicts that the processing effort incurred by a unit is monotonically related to its surprisal. In practice, empirical tests of surprisal theory rely on large pretrained language models as proxies for $\pHuman$ \citep{wilcox-etal-2023-testing, oh-schuler-2023-surprisal, shain2024logrithmic, kuribayashi-etal-2024-psychometric}. 
Additionally, evaluating the theory requires a \defn{linking hypothesis}, i.e., a specification of how surprisal maps onto an observable dependent variable such as reading time. Much attention has been devoted to the functional form of this mapping, whether processing effort scales linearly \citep{smith2013, shain2024logrithmic, wilcox-etal-2023-testing}, sublinearly \citep{brothers2021Word}, or superlinearly \citep{hoover2023Plausibility} with surprisal, while \citet{xu-etal-2023-linearity} find that the shape depends on the language model. However, as we argue in this paper, the choice of units and the aggregation strategy are equally consequential.\looseness=-1

The training set $\{\utterance^{n}\}_{n=1}^N$ consists of $N$ distinct utterances.
We write $T_n$ for the number of units in the $n^{\text{th}}$ utterance. 
For each $\unit^{n}_t$ given preceding context $\unitseq^{n}_{<t}$, we measure a reading time $\psychometric_\pi(\unit^{n}_t, \unitseq^{n}_{<t})$ from one of the $P$ participants.
Because fixation durations are strictly positive and right-skewed, we model reading times with a log-normal generalized additive mixed model \citep[GAMM;][]{wood-2017-gam}.
At position $t \in [T_n]$ of utterance $n$, let $\predvec_t^{n} = (x_{1,t}^{n}, \ldots, x_{J,t}^{n})^\top$ denote the vector of $J$ predictors. We model
\begin{equation}\label{eq:lognormal}
\log \psychometric_\pi(\unit^{n}_t, \unitseq^{n}_{<t}) = \mu_\pi(\unit^{n}_t, \unitseq^{n}_{<t}) + \epsilon,
\end{equation}
where $\epsilon\sim \mathcal{N}(0, \sigma^2)$ is Gaussian noise, $\sigma^2$ is the residual variance, and the log-mean is given by
\begin{equation}\label{eq:gamm}
\mu_\pi(\unit^{n}_t, \unitseq^{n}_{<t})  = \sum_{j=1}^{J} f_j(x_{j,t}^{n}) + z_\pi(\predvec_t^{n}).
\end{equation}
Each $f_j$ is a penalized smooth function of the $j^{\text{th}}$ component $x_{j,t}^{n}$ of $\predvec_t^{n}$, so that the relationship between each predictor and reading time is learned nonparametrically.
See \cref{sec:baseline} for a discussion of the predictors.
The term $z_{\pi}(\predvec_t^{n})$ captures participant-level random effects: a random intercept and by-participant random slopes for each predictor; see \cref{sec:gamm-spec} for the full specification.

\subsection{Baseline Predictors}
\label{sec:baseline}
A unit's \defn{length} and \defn{frequency} are standard baseline controls in eye-tracking regressions \citep{demberg-2008, smith2013, goodkind-bicknell-2018-predictive}: short, high-frequency units are more likely to be skipped, and when fixated, receive shorter fixation durations than longer, lower-frequency ones \citep{rayner_raney_1996_word_frequency, kliegl2004length}. Accordingly, evaluations of contextual surprisal typically include both as baseline predictors \citep{wilcox-etal-2023-testing, opedal-etal-2024-role, kuribayashi-etal-2024-psychometric}, with frequency operationalized as \defn{unigram surprisal}.\footnote{See \citet{shain-2019-large, shain-2024-word} and \citet{opedal-etal-2024-role} for discussion of unigram surprisal in analyzing reading times.} Prior work either computes word frequencies on held-out data \citep{pimentel-meister-2024-compute} or relies on precompiled lexical resources \citep[e.g.,][]{wilcox-etal-2023-testing, opedal-etal-2024-role, re2025spatiotemporal}, using toolkits such as \citet{robyn_speer_2022_7199437}. However, these convenient methods introduce two nontrivial mismatches. First, \citet{robyn_speer_2022_7199437} conflates distinct orthographic forms by stripping punctuation; for instance, it assigns \unittext{and} the same probability as \unittext{and$\cdot$,}. Moreover, the resulting unigram distribution is not aligned with the language model used to derive contextual surprisal, complicating comparisons between frequency and contextual predictability. We thus follow \citet{hopton2026unigram} and estimate unigram surprisal directly from the language model: we sample text from the LM, process each sample through the transduced LM to obtain per-unit conditional probabilities, and average these over all positions. The resulting unigram distribution is consistent with the model's own distribution and is naturally defined for every unit in our inventories; see \cref{app:experiments} for additional details. We include unigram surprisal estimated in this way as a baseline predictor in all analyses. Finally, to account for spillover effects we include the predictors for the preceding unit as controls \citep{rayner-1983}.\looseness=-1

\subsection{Predictive Power}
We evaluate the contribution of surprisal by comparing two instances of the model in \cref{eq:lognormal,eq:gamm}: a \defn{baseline model} $\widetilde{\varphi}$, in which the log-mean $\widetilde{\mu}_\pi(\unit^{n}_t, \unitseq^{n}_{<t})$ depends only on control predictors (unit length, unigram surprisal, and their spillover lags; see \cref{sec:baseline}),
and a \defn{target model} $\varphi$ that additionally includes contextual surprisal and its spillover lags (see \cref{sec:gamm-spec} for the full specification). We fit both models on the $N$ training utterances and evaluate them on a held-out test set of $M$ utterances, where utterance $m$ spans $T_m$ positions. Treating the end-of-sequence symbol $\eos$ as the $(T_m+1)^{\text{th}}$ unit, let $\mathcal{I} \defeq \{(m,t) : m \in [M],\, 1 \le t \le T_m + 1\}$ denote the set of held-out (utterance, position) pairs. For brevity, we write $\psychometric_t^{m} \defeq \psychometric_\pi(\unit^{m}_t, \unitseq^{m}_{<t})$ and $\mu_t^{m} \defeq \mu_\pi(\unit^{m}_t, \unitseq^{m}_{<t})$ (and $\widetilde{\mu}_t^{m}$ for the baseline). Writing the log-normal density from \cref{eq:lognormal} as
\begin{align}\label{eq:lognormal-density}
&\varphi\mleft(\psychometric_t^{m} \mid \mu_t^{m}, \sigma^2\mright) \defeq \nonumber \\ &\quad\frac{1}{\psychometric_t^{m}\,\sigma\sqrt{2\pi}}\, \exp\!\left(-\frac{\left(\log \psychometric_t^{m} - \mu_t^{m}\right)^2}{2\sigma^2}\right),
\end{align}
we measure the mean per-observation improvement in held-out log-likelihood, which is defined as 
\begin{equation}\label{eqn:delta-llh}
\dll \defeq \frac{1}{|\mathcal{I}|} \sum_{(m,t) \in \mathcal{I}} \log \frac{\varphi(\psychometric_{t}^{m} \mid \mu_{t}^{m},\, \sigma^2)}{\widetilde{\varphi}(\psychometric_{t}^{m} \mid \widetilde{\mu}_{t}^{m},\, \widetilde{\sigma}^{2})},
\end{equation}
where $\mu_t^{m}$ and $\widetilde{\mu}_{t}^{m}$ are the predicted log-means from the target and baseline models at position $t$ of held-out utterance $m$, and $\sigma$ and $\widetilde{\sigma}$ are the corresponding residual standard deviations, both estimated on the training set. A positive $\dll$ indicates that contextual surprisal captures variance in reading times beyond the baseline controls.

\section{Experiments}
\label{sec:experiments}
We now evaluate the predictive power of surprisal theory under four unit inventories.

\subsection{Unit Inventories}
\label{sec:unit-inventories}
\paragraph{Tokens.}
The simplest option is to use the model's native tokens as units, i.e., $\unitAlphabet = \srcAlphabet$. This is a natural choice when the objective is to characterize the model itself, e.g., when comparing the impact of token granularity on surprisal \citep{oh-schuler-2025-impact}, or to evaluate the cognitive plausibility of token-like representations \citep{beinborn-pinter-2023-analyzing, nair-resnik-2023-words}.\footnote{BPE can be encoded as a finite state machine \citep{Berglund-2023-formalizing, berglund2024bpe}.} 
However, in other experimental paradigms, model tokens are a poor fit, as they rarely align with linguistically meaningful units \citep{Church_2020, hofmann-etal-2021-superbizarre, nair-resnik-2023-words}. Another limitation of tokens is their coarse, model-dependent granularity, which can obscure effects that are naturally defined at finer spatial scales in the stimulus \citep{rayner1975parafoveal, Schotter2012}; see, for example, \Cref{ex:parafoveal_mismatch}.

\paragraph{Characters.}
At the other extreme, individual characters can constitute units. Character-level surprisal may be useful when the ROIs are sublexical, such as punctuation \citep{Rayner01112000, hill-2000, hirotani-2006}, morphological structure \citep{nair-resnik-2023-words}, the first few characters to predict the skip rate of a word \citep{rayner1982availability, blanchard-etal-1989-acquisition}, or as sublexical information used to improve surprisal estimates for larger units \citep{oh-etal-2021-surprisal}. 
We study character-level units in their own right, not merely as building blocks for computing ROI-level predictors.

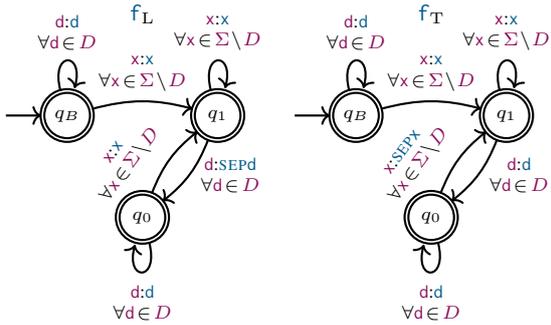
\begin{figure}[t]
\centering
\begin{tikzpicture}[
    auto, thick, ->,
    font=\scriptsize,
    state/.style={circle, draw, minimum size=6.5mm, inner sep=1pt},
    accepting/.style={double},
    initial text={}
]

\begin{scope}[xshift=0cm, yscale=0.9, xscale=0.9]
\node at (1.1,1.5) {\footnotesize $\STLEAD$};

\node[state, accepting, initial] (Lb) at (0,0) {$q_B$};
\node[state, accepting]          (L1) at (2.2,0) {$q_1$};
\node[state, accepting]          (L0) at (1.1,-1.5) {$q_0$};

\draw (Lb) edge[loop above]
  node[above, align=center] {\srcfont{d}:\tgtfont{d}\\ $\forall\srcfont{d}\!\in\!\delims$} ();

\draw (Lb) to[bend left=15]
  node[above, align=center] {\srcfont{x}:\tgtfont{x}\\ $\forall\srcfont{x}\!\in\!\srcAlphabet\!\setminus\!\delims$} (L1);

\draw (L1) edge[loop above]
  node[above, align=center] {\srcfont{x}:\tgtfont{x}\\ $\forall\srcfont{x}\!\in\!\srcAlphabet\!\setminus\!\delims$} ();

\draw (L1) to[bend left=15]
  node[right, align=center] {\srcfont{d}:\SEP\tgtfont{d}\\ $\forall\srcfont{d}\!\in\!\delims$} (L0);

\draw (L0) edge[loop below]
  node[below, align=center] {\srcfont{d}:\tgtfont{d}\\ $\forall\srcfont{d}\!\in\!\delims$} ();

\draw (L0) to[bend left=15]
  node[left, sloped, align=center, yshift=15pt, xshift=10pt]
  {\srcfont{x}:\tgtfont{x}\\ $\forall\srcfont{x}\!\in\!\srcAlphabet\!\setminus\!\delims$} (L1);
\end{scope}

\begin{scope}[xshift=3.8cm, yscale=0.9, xscale=0.9]
\node at (1.1,1.5) {\footnotesize $\STTRAIL$};

\node[state, accepting, initial] (Rb) at (0,0) {$q_B$};
\node[state, accepting]          (R1) at (2.2,0) {$q_1$};
\node[state, accepting]          (R0) at (1.1,-1.5) {$q_0$};

\draw (Rb) edge[loop above]
  node[above, align=center] {\srcfont{d}:\tgtfont{d}\\ $\forall\srcfont{d}\!\in\!\delims$} ();

\draw (Rb) to[bend left=15]
  node[above, align=center] {\srcfont{x}:\tgtfont{x}\\ $\forall\srcfont{x}\!\in\!\srcAlphabet\!\setminus\!\delims$} (R1);

\draw (R1) edge[loop above]
  node[above, align=center] {\srcfont{x}:\tgtfont{x}\\ $\forall\srcfont{x}\!\in\!\srcAlphabet\!\setminus\!\delims$} ();

\draw (R1) to[bend left=15]
  node[right, align=center] {\srcfont{d}:\tgtfont{d}\\ $\forall\srcfont{d}\!\in\!\delims$} (R0);

\draw (R0) edge[loop below]
  node[below, align=center] {\srcfont{d}:\tgtfont{d}\\ $\forall\srcfont{d}\!\in\!\delims$} ();

\draw (R0) to[bend left=15]
  node[left, sloped, align=center, yshift=15pt, xshift=10pt]
  {\srcfont{x}:\SEP\tgtfont{x}\\ $\forall\srcfont{x}\!\in\!\srcAlphabet\!\setminus\!\delims$} (R1);
\end{scope}

\end{tikzpicture}
\caption{Two delimiter-insertion transducers. Left: $\STLEAD$ inserts $\SEP$ before the first delimiter following each unit. Right: $\STTRAIL$ inserts $\SEP$ after that delimiter. These distinguish leading- and trailing-whitespace decoding \citep{oh-schuler-2024-leading, pimentel-meister-2024-compute}; see \cref{app:leading-vs-trailing} for discussion.\vspace{-15pt}
}
\label{fig:delimiter_transducers}
\end{figure}

\paragraph{Acontextual Words.}
A common choice is to define units using explicit orthographic rules. For instance, one could define a word-like notion by choosing the delimiter set $\srcAlphabetOne$ from the partition $\srcAlphabetOne \sqcup \srcAlphabetTwo$ of $\srcAlphabet$ introduced in \cref{sec:lost-in-whitespace}, such as whitespace \citep{wilcox-etal-2023-testing, pimentel-meister-2024-compute}, and splitting on those delimiters.\footnote{A design choice here is whether you also delete the delimiters or leave them in the resulting units.}
As illustrated in \cref{fig:delimiter_transducers}, this kind of segmentation can be implemented with a finite transducer. However, this is a modeling convenience inherited from how popular eye-tracking corpora such as Dundee \citep{kennedy-etal-2003-dundee}, Provo \citep{luke-etal-2018-provo}, or MECO \citep{siegelman2022expanding} distribute their data, and is not meant to represent a linguistically adequate notion of a word.
Its simplicity is also its main limitation: splitting on delimiters cannot express context-dependent boundaries. Consider, for example, how punctuation is typically considered its own unit in a context such as \symboltext{long, tiring trip} but not in \symboltext{1,000}. The same holds for internal apostrophes in contractions or in abbreviations. In fact, several recent studies have excluded all words attached to punctuation \citep{nair-resnik-2023-words, klein-etal-2024-effect, clark-2025}. However, research has long reported the systematic effects of punctuation on reading behavior (e.g., pauses and longer first pass times around commas) \citep{Rayner01112000, hill-2000, hirotani-2006}. Handling such cases requires contextual segmentation rules.

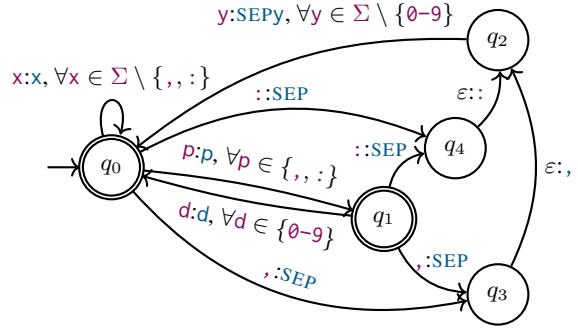
\begin{figure}
\centering
\begin{tikzpicture}[node distance=1cm,
    auto,
    thick,
    state/.style={circle, draw, minimum size=0.8cm},
    accepting/.style={double}, initial text={}
    ]
\footnotesize
\node[state, initial, accepting] (q0) {$q_0$};
\node[state, accepting] (q1) [below right=1mm and 30mm of q0] {$q_1$};
\node[state] (q2) [above right=11mm and 45mm of q0] {$q_2$};
\node[state] (q3) [below right=11mm and 45mm of q0] {$q_3$};
\node[state] (q4) [above right=5mm of q1] {$q_4$};

\draw[->] (q0) edge[bend right=30]
node[above, sloped]
{\srcfont{,}:$\SEP$} (q3);

\draw[->] (q0) edge[bend left=5]
node[above, sloped, xshift=0.1cm]
{\srcfont{p}:\tgtfont{p}, $\forall \srcfont{p}\in\{\srcfont{\texttt{,}},\srcfont{\texttt{:}}\}$} (q1);

\draw[->] (q1) edge[bend left=5]
node[below, sloped, xshift=0.2cm]
{\srcfont{d}:\tgtfont{d}, $\forall \srcfont{d}\in\{\srcfont{\texttt{0--9}}\}$} (q0);

\draw[->] (q1) edge[bend right=30]
node[above, xshift=0.2cm]
{$\srcfont{,}\text{:}\SEP$} (q3);

\draw[->] (q3) edge[bend right=25]
node[right]
{$\varepsilon$:$\tgtfont{,}$} (q2);

\draw[->] (q1) edge[bend left=35]
node[left, yshift=0.2cm, xshift=0.2cm]
{$\srcfont{:}\text{:}\SEP$} (q4);

\draw[->] (q0) edge[bend left=25]
node[above]
{$\srcfont{:}\text{:}\SEP$} (q4);

\draw[->] (q4) edge[bend right=25]
node[left, yshift=0.1cm]
{$\varepsilon$:$\tgtfont{:}$} (q2);

\draw[->] (q2) edge[bend right=20]
node[above, xshift=0.6cm, yshift=0.3cm]
{\srcfont{y}:$\SEP$\tgtfont{y}, $\forall \srcfont{y}\in\srcAlphabet\setminus\{\srcfont{\texttt{0--9}}\}$} (q0);

\draw[->] (q0) edge[loop above]
node {\srcfont{x}:\tgtfont{x}, $\forall \srcfont{x}\in\srcAlphabet\setminus\{\srcfont{\texttt{,}},\srcfont{\texttt{:}}\}$} ();
\end{tikzpicture}
\caption{A rule from $\STPTB$ showing contextual segmentation: a comma or colon is split off as its own unit (surrounded by $\SEP$s) only when the \emph{following} symbol is not a digit, e.g. \symboltext{end, he} is split into three units, while \symboltext{1,000} remains one. Adapted from \citet{snbjarnarson2026transducing}. \vspace{-15pt}
}
\label{fig:contextual_rule_main}
\end{figure}

\paragraph{Contextual Words.}
The acontextual assumption, i.e., that a fixed partition of $\srcAlphabet$ suffices to determine unit boundaries, is linguistically inadequate, because the same symbol can mark a boundary in one context but not another.
Thus, instead of defining units via an explicit set of delimiters, it is common practice to use contextual segmentation rules, such as the Penn Treebank guidelines \citep{marcus-etal-1993-building}, or Universal Dependencies \citep{nivre-etal-2017-universal}. 
Such units are linguistically informed, but introduce an additional challenge with token-level language models, where tokens are not compatible with the resulting units. Here we follow \citet{snbjarnarson2026transducing} and encode each rule\footnote{For an overview of the individual tokenization rules, see \href{https://www.nltk.org/_modules/nltk/tokenize/treebank.html\#TreebankWordTokenizer}{\texttt{TreebankWordTokenizer}}.} as a finite transducer and then compose them left to right to obtain $\STPTB$; see \cref{sec:ptb_construction} for additional details. 
In contrast to the acontextual segmentation in \cref{fig:delimiter_transducers}, this transducer determines word boundaries using contextual information. Consider, for example, the rule in \cref{fig:contextual_rule_main}, which inserts a separator $\SEP$ before a comma ($\srcfont{,}$) or a colon ($\srcfont{:}$) only if the following symbol is not a digit.\looseness=-1

\begin{figure*}
    \centering
    \includegraphics[width=\textwidth]{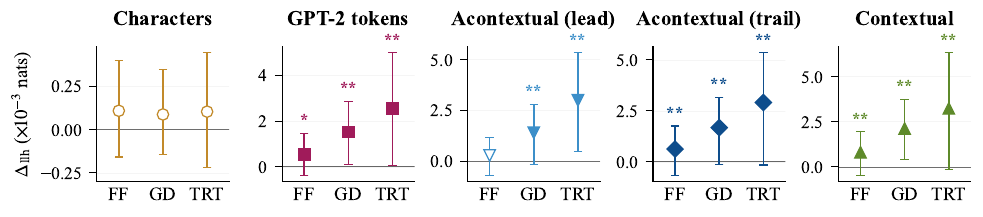}
    \caption{Per-observation $\dll$ ($\times 10^{-3}$ nats) for each unit inventory across reading-time measures (FF: first fixation, GD: gaze duration, TRT: total reading time). Points and whiskers show the mean and 95\% trial-level bootstrap CI from leave-one-out cross-validation by trial. Significance is assessed via a paired permutation test ($^{*}$\,$\significance < 0.05$; $^{**}$\,$\significance < 0.01$). Filled markers denote significant effects. Note that $y$-axis scales differ across panels: log-likelihoods are not comparable across inventories because the number and granularity of observations differ.
    }
    \label{fig:gamm_results}
\end{figure*}

\subsection{(Re-)processing the MECO Corpus}

To obtain fixation data for each unit inventory discussed in \Cref{sec:unit-inventories}, we process the raw fixation data from the MECO dataset \citep{siegelman2022expanding}, which contains scanpaths from 46 readers recorded while reading 12 short excerpts drawn from Wikipedia articles.
We use the English portion of the dataset.
Following \citet{re2025spatiotemporal}, we first obtain the unprocessed fixation data and use the predefined bounding boxes to match fixations with individual characters. We then tokenize the raw text and aggregate the fixation durations within the boundaries of each unit, to obtain three commonly used \citep{rayner-1998} reading time measurements $\psychometric(\unit_t, \unitseq_{<t})$: \defn{first-fixation duration}, the duration of the first fixation on unit $\unit_t$; \defn{gaze duration}, the sum of all first-pass fixations on $\unit_t$; and \defn{total reading time}, the sum of all fixations on $\unit_t$, including any refixations. Fixations on whitespace characters are credited to whichever unit the transducer assigns that whitespace to.
We exclude unfixated units and retain per-reader observations for use with mixed-effects models (\cref{sec:gamm-spec}). For the character inventory, we additionally exclude observations whose surprisal is exactly zero, which arise at sub-token byte positions. In \cref{app:unit-visualizations}, we visualize the resulting units and fixations; \cref{tab:units_stats} reports the observation counts at each step of the pipeline.\looseness=-1

\subsection{Estimating Surprisal}
We use GPT-2 Small \citep{radford2019language} as our symbol language model $\pSource$, following previous work \citep{oh-schuler-2023-surprisal}. For the token inventory, surprisal is read directly from $\pSource$. For all other inventories, we compose $\pSource$ with the appropriate finite transducer; see \cref{app:experiments} for details. 

\subsection{Analysis}
We fit the log-mean of \cref{eq:lognormal} for both baseline and target models as a generalized additive mixed model \citep[GAMM;][]{wood-2017-gam}, which estimates the contribution of each predictor through a penalized smooth function. The residual standard deviations ($\sigma$ and $\widetilde{\sigma}$) are estimated on the log scale of the training set. To assess generalization, we perform leave-one-out cross-validation by trial (12~folds): in each fold, we fit both models on 11 trials and compute $\dll$ (\cref{eqn:delta-llh}) on the held-out trial. We report the per-observation $\dll$ with 95\% confidence intervals obtained by trial-level bootstrap (1000~iterations). Significance of $\dll$ is assessed by a one-sided paired sign-flip permutation test over held-out log-likelihoods; see \cref{sec:gamm-spec} for details.\looseness=-1

\subsection{Results}\label{sec:results}

\cref{fig:gamm_results} shows our results across unit inventories and reading-time measures; detailed results are given in \cref{tab:gamm_all}.
Adding surprisal as a predictor yields significant improvements in $\dll$ over the baseline for word-like inventories on later reading-time measures (gaze duration and total reading time; all $\significance < 0.001$, paired permutation test), in line with previous findings \citep{goodkind-bicknell-2018-predictive, wilcox-etal-2023-testing}. For first-fixation duration results are significant for tokens, contextual words, and acontextual words under trailing-whitespace attribution, but not under leading-whitespace attribution. The character inventory shows no significant $\dll$ on any measure, and its magnitudes are markedly smaller, as single characters are rarely fixated individually. \looseness=-1

A central lesson is that changing the unit of analysis changes the regression problem itself: different inventories induce different observations and controls (length, spillover, unigram surprisal), so absolute log-likelihoods and $\dll$ values are not directly comparable across inventories (\cref{tab:units_stats}).
This lesson matters beyond the inventories considered here: existing work has evaluated surprisal at granularities from morphemes \citep{nair-resnik-2023-words} and phonemes \citep{brodbeck-2022-parallel, tezcan-2023-phoneme, sohoglu-2024-syllables} to sentences \citep{lau-2017-sentences, giulianelli2023information} and discourse units \citep{tsipidi-etal-2024-surprise, tsipidi-etal-2025-harmonic}. More broadly, the framework applies wherever the unit of analysis must be specified, whether the dependent variable is reading time, neural signals \citep{frank-etal-2013-word, frank2015ERP, kuribayashi2025Large}, or a modified predictive distribution such as lossy-context surprisal \citep{futrell2020Lossy} or syntactic surprisal \citep{demberg-2008, arehalli-etal-2022-syntactic}.\looseness=-1

\section{Conclusion}
We make a simple point: next-unit contextual surprisal is only well-defined relative to a choice of unit inventory. Yet existing work often inherits the language model's tokenizer. 
We present a formalism that makes this choice explicit and returns it to the modeler.
We describe a principled way to derive unit-level surprisal from token-level language models. Because the choice of unit reshapes the regression problem and its baseline predictors, we argue it should be treated as a first-class modeling decision.
We encourage future work to actively select the units most appropriate for their analysis.

\section*{Limitations}
This study is primarily methodological, discussing the appropriate use of units and ROIs in surprisal theory and is therefore limited in scope. 
\paragraph{Unit Inventories and ROIs.}
Our empirical evaluation is restricted to tokens, characters, and two families of word-like segmentations (contextual and acontextual). Beyond these units, researchers have studied several other unit inventories and ROIs, such as discourse units (e.g., clauses or elementary discourse units). Our framework naturally extends to such inventories, and we leave their empirical investigation to future work. 

\paragraph{Language and Model Coverage.}
Our empirical results are restricted to English: what counts as a word varies with orthography and linguistic traditions, and many languages require different segmentation rules than those standard in English \citep{nivre-etal-2017-universal}. In addition, our analysis evaluates only GPT-2 Small on the MECO dataset, using GAMMs to predict reading times. Future studies could therefore broaden the empirical analyses to evaluate unit inventories and ROI choices across models and datasets.

\paragraph{Data Requirements.}
Our analysis is contingent on access to raw fixation data. Many published reading-time corpora distribute only pre-aggregated word-level reading times, and self-paced reading datasets are inherently bound to a fixed segmentation. In such cases, fixations cannot be re-aggregated to alternative unit boundaries, and our approach can only be applied if the chosen unit inventory is compatible with the corpus's pre-existing segmentation.

\paragraph{Computational Costs.}
Computing surprisal under the transduced language model requires marginalizing over all source strings that map to a given output, which incurs computational overhead that depends on the transducer and the source language model. In our experiments, we use the beam-search approximations described in \citet{snbjarnarson2026transducing}. We argue here that the resulting throughput (\cref{tab:throughput-surprisal}) is sufficient to make contextual surprisal estimation computationally feasible for typical psycholinguistic corpora. Estimating unigram surprisal is considerably more demanding and in practice requires parallelization across samples; see \cref{app:experiments} for details. A lighter-weight alternative is to estimate unigram probabilities by transducing the sampled text and directly counting unit occurrences, which sidesteps sequential per-boundary scoring of every candidate unit but can underestimate units that rarely appear in the samples; see \citet{hopton2026unigram}. Another alternative would be a hybrid approach that uses sample counts for frequent units and falls back to the conditional estimate for rare units.

\paragraph{Expressivity.}
Our framework inherits the expressivity limits of finite-state machinery: we assume the unit parser $\parsingf$ is deterministic, and that it is rational, i.e., realizable by a finite transducer. Phenomena beyond this scope, such as genuinely ambiguous parsing and non-rational transformations such as those requiring context-free structure, fall outside the current framework and are left to future work.\looseness=-1

\section*{Ethical Considerations}
This work is a conceptual study about the role of units in psycholinguistic theory.
The datasets we use are public and released with the consent of all participants. All personally identifiable information had been removed prior to our use of the data. As such, we do not see any ethical problems with this work.

\section*{Acknowledgments}
The authors would like to thank Andreas Opedal, Francesco Ignazio Re, Jacob Hoover Vigly, Zach Hopton, Eleftheria Tsipidi, Mario Giulianelli, and Thomas Hikaru Clark for their valuable feedback and helpful discussions. VS is supported by the Pioneer Centre for AI, DNRF grant number P1. 
We used generative AI to assist with writing and with debugging code. The code and the writing were carefully reviewed and verified by the authors, who take full responsibility for the content of this paper. 

\bibliography{custom}
\newpage
\onecolumn
\appendix

\startappendixtoc

\appendixtableofcontents
\clearpage

\section{Notation Glossary}
\label{app:notation}

\begin{table}[!h]
\centering
\begin{tabular}{@{}lp{12cm}@{}}
\toprule
\textbf{Symbol} & \textbf{Meaning} \\
\midrule
\multicolumn{2}{@{}l}{\textit{Alphabets and Strings}} \\
\midrule
$\srcAlphabet$ & Symbol alphabet (characters or tokens). \\
$\tgtAlphabet$ & Output alphabet of the transducer. \\
$\sepAlphabet$ & Finite alphabet over which units are strings; $\unitAlphabet \subseteq \kleene{\sepAlphabet}$ and $\tgtAlphabet = \sepAlphabet \sqcup \{\SEP\}$ (\cref{sec:regularity}). \\
$\unitAlphabet$ & Unit inventory chosen by the modeler (countable, possibly infinite). \\
$\SEP$ & Distinguished separator symbol marking unit boundaries; $\SEP \notin \sepAlphabet$. \\
$\eos$ & End-of-sequence symbol; $\eos \notin \Sigma$. \\
$\kleene{\Sigma}$ & Kleene closure: set of all finite strings over alphabet $\Sigma$, including $\varepsilon$. \\
$\kleeneplus{\Sigma}$ & Non-empty strings: $\kleene{\Sigma} \setminus \{\varepsilon\}$. \\
\midrule
\multicolumn{2}{@{}l}{\textit{Units and Contexts}} \\
\midrule
$T$ & Length of a string or utterance (number of symbols or units). \\
$\unit_t$ & The $t$-th unit in an utterance; $\unit_t \in \unitAlphabet$. \\
$\utterance$ & Utterance: sequence of units, $\utterance = \unit_1 \cdots \unit_T \in \unitStrings$. \\
$\context$ & Preceding-unit context. \\
$\utterance_{[i,j)}$ & Region of interest (ROI): subspan $\unit_i \cdots \unit_{j-1}$. \\
\midrule
\multicolumn{2}{@{}l}{\textit{Maps and Transducers}} \\
\midrule
$\parsingf$   & Unit parser (stochastic map $\parsingf \colon \srcStrings \rightsquigarrow \unitStrings$); assumed deterministic ($\srcStrings \to \unitStrings$) in this paper. \\
$\realization$ & Realization: a relation $\realization \subseteq \unitStrings \times \srcStrings$ mapping unit strings to symbol strings. \\
$\transFn$ & String-to-string relation $\transFn \subseteq \srcStrings \times \kleene{\tgtAlphabet}$. \\
$\transST$ & Finite transducer with input alphabet $\srcAlphabet$ and output alphabet $\tgtAlphabet$. \\
$\delims$ & Set of delimiter symbols; $\delims \subseteq \srcAlphabet$. \\
$\STLEAD,\;\STTRAIL$ & Acontextual (delimiter-based) FSTs: leading and trailing $\SEP$ insertion. \\
$\STPTB$ & Contextual FST implementing Penn Treebank segmentation rules. \\
$\transH$ & Homomorphism $\unitAlphabet \to \kleene{\sepAlphabet}\,\SEP$ appending $\SEP$ to each unit's underlying string; extended to $\unitStrings$ by concatenation (\cref{eq:unit-homomorphism}). \\
$\transHinv$           & Inverse of $\transH$: splits on $\SEP$ and maps each segment to its unit (\cref{sec:transduced-lms}). \\
\midrule
\multicolumn{2}{@{}l}{\textit{Probability and Surprisal}} \\
\midrule
$\pSource$ & Source/token-level language model (probability distribution) over $\srcStrings$. \\
$\pTarget$ & Transduced language model over $\kleene{\tgtAlphabet}$, defined as $\pSource \circ \transST$. \\
$\pU$ & Unit-level language model over $\unitStrings$. \\
$\pHuman$ & Implicit human language model assumed by surprisal theory. \\
$\surprisal$ & Surprisal of a unit in context: $-\log \prefixU(\unit_t \mid \context)$. \\
\midrule
\multicolumn{2}{@{}l}{\textit{Reading-time Analysis}} \\
\midrule
$\psychometric_\pi(\unit^n_t, \unitseq^n_{<t})$      & Reading-time measurement (first-fixation, gaze, or total) for unit $\unit^n_t$ from participant $\pi$. \\
$\predvec^n_t$ & Predictor vector $(x_{1,t}^n, \ldots, x_{J,t}^n)^\top$ at position $t$ of utterance $n$.\\
$N,\;n$ & Number of training utterances; index over training utterances. \\
$M,\;m$ & Number of held-out (test) utterances; index over test utterances. \\
$\llb,\;\lltgt$ & Mean per-observation held-out log-likelihood of baseline / target GAMM. \\
$\dll$ & Improvement in held-out log-likelihood: $\lltgt - \llb$. \\
\bottomrule
\end{tabular}
\caption{Notation used throughout the paper.}
\label{tab:notation}
\end{table}
\clearpage

\section{Prefix-Freeness of $\transH$}
\label{app:trailing-h}

In \cref{eq:unit-homomorphism} we place $\SEP$ at the \emph{right} edge of each unit, i.e., $\SEP$ marks a unit's completion, so that $\transH(\unit) \defeq \sepStr_{\unit}\,\SEP$. One could equivalently consider the mirror convention
\begin{equation*}
\transH_{\mathrm{l}}(\unit) \defeq \SEP\,\sepStr_{\unit},
\end{equation*}
in which $\SEP$ instead marks a unit's \emph{onset}. Both are monoid homomorphisms $\unitStrings \to \kleene{\tgtAlphabet}$ whose images $\transH(\unitAlphabet)$ and $\transH_{\mathrm{l}}(\unitAlphabet)$ are regular and related by a 1-symbol shift of $\SEP$. The pushforward identity \cref{eq:next-unit}, however, is an equality only under the completion convention. The reason, as we show below, is that the completion convention makes $\transH(\unitAlphabet)$ a prefix-free code, whereas the onset convention does not.

\paragraph{Reduction.} Using $\transH(\utterance\c\unit) = \transH(\utterance)\c\transH(\unit)$, \cref{eq:next-unit} reduces to the prefix identity
\begin{equation}\label{eq:biconditional-app}
\parsingf(\srcStr) \succeq \unitseq \iff \transH(\parsingf(\srcStr)) \succeq \transH(\unitseq),
\end{equation}
for all $\srcStr \in \srcStrings$ and $\unitseq \in \unitStrings$.
The $(\Rightarrow)$ direction holds for any monoid homomorphism; only the $(\Leftarrow)$ direction depends on the placement of $\SEP$.

\paragraph{Completion ($\SEP$ trailing).} Writing $\unitseq = \unit_1\c\cdots\c\unit_k$ and $\parsingf(\srcStr) = \unit'_1\c\cdots\c\unit'_T$, we have
\begin{align*}
\transH(\unitseq) &= \sepStr_{\unit_1}\,\SEP\,\cdots\,\sepStr_{\unit_k}\,\SEP, \\
\transH(\parsingf(\srcStr)) &= \sepStr_{\unit'_1}\,\SEP\,\cdots\,\sepStr_{\unit'_T}\,\SEP.
\end{align*}
Since $\SEP \notin \sepAlphabet$, every $\SEP$ in $\transH(\parsingf(\srcStr))$ sits at a block boundary. Hence, if $\transH(\parsingf(\srcStr)) \succeq \transH(\unitseq)$, then the \emph{final} $\SEP$ of $\transH(\unitseq)$ must coincide with the block boundary after $\sepStr_{\unit'_k}$. Matching back block-by-block forces $\unit'_i = \unit_i$ for $i \leq k$, and thus $\parsingf(\srcStr) \succeq \unitseq$.

\paragraph{Onset ($\SEP$ leading).} Under $\transH_{\mathrm{l}}$, the image $\transH_{\mathrm{l}}(\unitseq)$ ends with the bare bytes $\sepStr_{\unit_k}$: the $\SEP$ that would close $\unit_k$ is absorbed into the onset of $\unit_{k+1}$ and is therefore absent from $\transH_{\mathrm{l}}(\unitseq)$. A byte-prefix match then admits any $\unit' \in \unitAlphabet$ whose underlying string begins with $\sepStr_{\unit_k}$, not only $\unit_k$ itself.

\paragraph{Counter-example.} Let $\srcAlphabet = \{\symboltext{a}, \symboltext{b}\}$, $\sepAlphabet = \{\tgtfont{a}, \tgtfont{b}\}$, and $\tgtAlphabet = \sepAlphabet \sqcup \{\SEP\}$. Take $\unitAlphabet = \{\unit_a, \unit_{ab}\} \subset \kleene{\sepAlphabet}$ with underlying strings $\sepStr_{\unit_a} = \tgtfont{a}$ and $\sepStr_{\unit_{ab}} = \tgtfont{ab}$, a deterministic $\parsingf$ satisfying $\parsingf(\symboltext{a}) = \unit_a$ and $\parsingf(\symboltext{ab}) = \unit_{ab}$, and source distribution $\pSource(\symboltext{a}) = \pSource(\symboltext{ab}) = \tfrac{1}{2}$, so that $\prefixU(\unit_a) = \tfrac{1}{2}$. Under the onset convention, the transducer produces target strings $\SEP\,\tgtfont{a}$ and $\SEP\,\tgtfont{ab}$, both of which have $\transH_{\mathrm{l}}(\unit_a) = \SEP\,\tgtfont{a}$ as a byte prefix; scoring with the transduced LM therefore gives $\prefixTarget(\transH_{\mathrm{l}}(\unit_a)) = 1 \neq \tfrac{1}{2}$. Under the completion convention, $\tgtfont{a}\,\SEP$ is not a byte prefix of $\tgtfont{ab}\,\SEP$, and $\prefixTarget(\tgtfont{a}\,\SEP) = \tfrac{1}{2}$ matches $\prefixU(\unit_a)$ exactly.

\paragraph{Practical consequence.} The trailing $\SEP$ inside each $\transH(\unit_k)$ pins the unit's right boundary in the target-byte prefix to a block boundary of the parse, ruling out parses in which $\unit_k$ has been silently extended into a longer unit sharing its byte prefix. Whitespace-delimited English inventories contain many such byte-prefix overlaps---\unittext{the}$\subset$\unittext{there}, \unittext{in}$\subset$\unittext{into}, \unittext{a}$\subset$\unittext{an}$\subset$\unittext{and}---so the onset convention would leak probability mass pervasively in practice.
\clearpage

\section{Transducers}
\label{ssec:transducers}

Here we provide additional details on the FSTs used in \cref{sec:experiments}. We implement all FSTs in Pynini \citep{gorman-2016-pynini}, a Python library for compiling and composing finite transducers that builds on OpenFST \citep{riley-etal-2009-openfst}. 

\subsection{Characters}
\begin{figure}[ht]
\centering
\centering
\begin{tikzpicture}[shorten >=1pt, on grid, thick, auto,
  node distance=18mm, initial text={}] 
\footnotesize
\node[state, initial, initial where=above, accepting] (q0) {$q_0$};

\node[state] (q1) [above right=8mm and 18mm of q0] {$q_1$};
\node[state] (q2) [right=26mm of q0] {$q_2$};
\node[state] (q3) [below right=8mm and 18mm of q0] {$q_3$};

\node[state] (q4) [above left=8mm and 18mm of q0] {$q_4$};
\node[state] (q5) [below left=8mm and 18mm of q0] {$q_5$};

\draw[->] (q0) to node[sloped, below] {\srcfont{\wsp{}cat}:\tgtfont{\wsp{}}} (q1);
\draw[->] (q1) to node {$\varepsilon$:\tgtfont{c}} (q2);
\draw[->] (q2) to node {$\varepsilon$:\tgtfont{a}} (q3);
\draw[->] (q3) to node[sloped, below]
  {$\varepsilon$:\tgtfont{t}} (q0);

\draw[->] (q0) to node[sloped, below] {\srcfont{Dog}:\tgtfont{D}} (q4);
\draw[->] (q4) to[bend right=20] node {$\varepsilon$:\tgtfont{o}} (q5);
\draw[->] (q5) to node[sloped, below]
  {$\varepsilon$:\tgtfont{g}} (q0);

\node (cdots) [below=12mm of q0] {$\cdots$};
\draw[->] (q0) to (cdots);
\end{tikzpicture}

\caption{A finite transducer for mapping a token-level LM to characters, illustrated with paths for \srcfont{\wsp{}cat} and \srcfont{Dog}. Adapted from \citet{snbjarnarson2026transducing}.}
\label{fig:subwordST}
\end{figure}
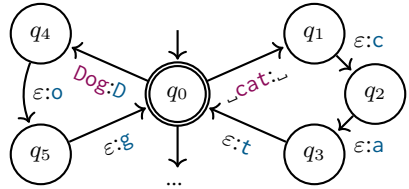

As shown by \citet{snbjarnarson2026transducing} (see \cref{fig:subwordST}), the transformation from tokens to characters can be encoded using an FST. In this work, we do not compile the FST for the experiments in \cref{app:experiments}; instead, we use the algorithms and implementations of \citet{pmlr-v267-vieira25a} to transform $\pSource$ into a character-level model.

\subsection{Acontextual Words}
\label{app:leading-vs-trailing}
Recent work argues that the common leading-whitespace convention of many BPE tokenizers---where the space preceding a word is bundled with the word's first token---induces a misallocation of surprisal, and that probability mass should instead be attributed as trailing whitespace, i.e., assigned to the preceding word \citep{oh-schuler-2024-leading, pimentel-meister-2024-compute}. At the same time, \citet{giulianelli-etal-2024-proper} argue that there is no universally correct convention and the attribution of whitespace should be chosen to match the experimental setup. 
Note that under the definition of \citet{oh-schuler-2024-leading} and \citet{pimentel-meister-2024-compute}, leading-versus-trailing whitespace decoding can be interpreted as an aggregation method that specifies how probability mass is redistributed across unit boundaries. Here, we express leading and trailing attribution as two finite transducers, $\STLEAD$ and $\STTRAIL$; see \cref{fig:delimiter_transducers}. This leading-versus-trailing choice concerns the \emph{delimiter byte's} unit membership (i.e., which unit the space belongs to) and is a separate axis from the trailing-$\SEP$ convention on $\transH$ itself; see \cref{app:trailing-h} for why the latter must be trailing regardless. In our experiments, we set the delimiter set to $\delims = \{\wspsym\}$, so whitespace is the sole signal of a unit boundary.

Note that a third variant, in which whitespace is absorbed into the separator ($q_1 \xrightarrow{\srcfont{d}:\SEP} q_0$) so that no unit ever contains a whitespace character, is easy to construct by replacing the delimiter arcs in \cref{fig:delimiter_transducers}.

\subsection{Contextual Words}
\label{sec:ptb_construction}
We represent contextual words using the rules described in the Penn Treebank annotation guidelines and encode each rule\footnote{\href{https://www.nltk.org/_modules/nltk/tokenize/treebank.html\#TreebankWordTokenizer}{\texttt{TreebankWordTokenizer}}.} as a small context-dependent string-rewrite transducer (see \cref{fig:contextual_rule_main} for an example of such a rule), using Pynini’s rewrite calculus (e.g., replace operations with explicit left/right contexts and boundary conditions). The full tokenizer is obtained by composing these rule transducers left-to-right, yielding a single transducer that maps input strings to their PTB-style tokenized form.

\subsection{Transducer Sizes}
\label{app:transducer-sizes}
\cref{tab:transducer-sizes} reports the number of states and arcs of each FST $\transST$ used in the experiments, together with its number of \emph{universal} states, i.e., those states, where the corresponding input-projected FSA accepts every symbol $\srcSym\in\srcAlphabet$ (see \citet{snbjarnarson2026transducing} for a detailed discussion). Universal states can be handled more efficiently in the algorithms provided by \citet{snbjarnarson2026transducing}, so transducers with a larger universal fraction yield a higher throughput (Syms/s). 

\begin{table*}[h]
\centering
\begin{minipage}[t]{0.48\textwidth}
\centering
\small
\setlength{\tabcolsep}{3pt}
\begin{tabular}{@{}lrrr@{}}
\toprule
Transducer & States & Arcs & Universal \\
\midrule
Acontextual (leading)  &   3 &    517 &  3 (all)   \\
Acontextual (trailing) & 258 &  1{,}029 & 258 (all)  \\
Contextual           & 361 & 35{,}210 &  78 (partial) \\
\bottomrule
\end{tabular}
\caption{Size of each finite transducer $\transST$ used in the experiments, together with the number of universal states \citep{snbjarnarson2026transducing}; ``all'' indicates that every state is universal.}
\label{tab:transducer-sizes}
\end{minipage}
\hfill
\begin{minipage}[t]{0.48\textwidth}
\centering
\small
\setlength{\tabcolsep}{4pt}
\begin{tabular}{@{}lrrr@{}}
\toprule
 & GPT-2 & Acontextual & Contextual \\
\midrule
GPT-2 tokens & --- & 70.8\% & 81.4\% \\
Acontextual & 70.9\% & --- & 85.0\% \\
Contextual & 84.2\% & 87.8\% & --- \\
\bottomrule
\end{tabular}
\caption{\textbf{Overlap with whitespace stripped}: percentage of units in the row inventory whose form---after removing any attributed whitespace---is also a unit in the column inventory. The acontextual leading and trailing variants yield identical stripped forms and are merged here.}
\label{tab:unit-overlap-stripped}
\end{minipage}
\end{table*}

The two delimiter-insertion transducers $\STLEAD$ and $\STTRAIL$ exhibit a size asymmetry despite being drawn as equivalent three-state machines in \cref{fig:delimiter_transducers}. This is because the compiled FSTs allow only one output symbol per arc, while the figure uses a shorthand to draw arcs with multiple-symbol outputs. Each such arc is compiled into a chain of two arcs through an auxiliary state, with one auxiliary per input-byte value. In $\STLEAD$ the multi-symbol output sits on the delimiter arc $q_1 \xrightarrow{\srcfont{d}:\SEP\tgtfont{d}} q_0$, so only $|\delims|$ auxiliary states are introduced; with $\delims = \{\wspsym\}$ the single resulting auxiliary is merged away by minimization, leaving three states. In $\STTRAIL$ the multi-symbol output sits on the arc $q_0 \xrightarrow{\srcfont{x}:\SEP\tgtfont{x}} q_1$, so $|\srcAlphabet \setminus \delims|$ auxiliaries are introduced; with a byte alphabet ($|\srcAlphabet|=256$) that is $255$ auxiliaries whose distinct output labels prevent merging, giving $3+255 = 258$ states.

\section{Dataset Details}
\label{app:datasets}
We reprocess the English portion of the MECO eye-tracking corpus \citep{siegelman2022expanding}, which contains scanpaths from 46 readers recorded while reading 12 short Wikipedia excerpts. \Cref{tab:units_stats} reports the number of observations at each stage of the analysis pipeline, from raw units to the final GAMM input, for each of the unit inventories.

\begin{table}[ht]
\centering
\small
\begin{tabular}{@{}l r r r@{}}
\toprule
\textbf{Inventory} & $n_{\text{units}}$ & $n_{\text{obs}}$ & $n_{\text{lag}}$ \\
\midrule
GPT-2 tokens         &  2{,}478 & 41{,}553 & 40{,}589 \\
Acontextual (leading)    &  2{,}095 & 40{,}254 & 39{,}290 \\
Acontextual (trailing)   &  2{,}095 & 40{,}731 & 39{,}767 \\
Contextual           &  2{,}264 & 34{,}478 & 33{,}514 \\
Characters           & 13{,}226 & 54{,}578 & 53{,}614 \\
\bottomrule
\end{tabular}
\caption{Pipeline observation counts per unit inventory. $n_{\text{units}}$: total units across all 12 trials. $n_{\text{obs}}$: per-reader observations after excluding zero reading times (unfixated units). $n_{\text{lag}}$: after dropping the first two units of each (reader, trial) pair, which lack values for the spillover lags. See \cref{tab:units_stats_gamm} for the final counts entering the GAMM.}
\label{tab:units_stats}
\end{table}

\subsection{Unit Overlap}

\begin{table}[h]
\centering
\small
\setlength{\tabcolsep}{3pt}
\begin{tabular}{@{}lrrrr@{}}
\toprule
 & GPT-2 & Acontextual (leading) & Acontextual (trailing) & Contextual \\
\midrule
GPT-2 tokens              & ---     & 70.5\%  & 0.0\%   & 2.1\% \\
Acontextual (leading)     & 71.0\%  & ---     & 0.0\%   & 0.7\% \\
Acontextual (trailing)    & 0.0\%   & 0.0\%   & ---     & 0.3\% \\
Contextual                & 2.2\%   & 0.7\%   & 0.3\%   & --- \\
\bottomrule
\end{tabular}
\caption{\textbf{Overlap with whitespace kept} (the unit text used by our GAMMs): percentage of units in the row inventory that appear \emph{verbatim}, including any attributed whitespace, as units in the column inventory.}
\label{tab:unit-overlap-ws}
\end{table}

With whitespace stripped (\cref{tab:unit-overlap-stripped}), the three word-like inventories share 71--88\% of their units pairwise. With whitespace kept (\cref{tab:unit-overlap-ws}), the two acontextual variants become nearly disjoint, and the contextual inventory, whose units never contain whitespace, is nearly disjoint from all three. GPT-2 tokens, which use leading-space by convention, overlap tightly with acontextual leading but not with acontextual trailing.

Looking at the whitespace-stripped differences: the 604 GPT-2 occurrences absent from the acontextual inventory are mainly punctuation marks that BPE splits off (126 commas, 96 periods) and BPE subword fragments (e.g., \srcfont{Jan}, \srcfont{us} from \textit{Janus}; conversely, the 325 acontextual occurrences absent from GPT-2 are words that BPE splits into multiple tokens (e.g., \textit{Janus}, \textit{thylacine}, \textit{performance-enhancing}). Acontextual and contextual words differ mainly on punctuation attachment: the contextual inventory splits off 129 commas and 11 periods that the acontextual inventory attaches to the preceding word (e.g., acontextual \unittext{organizations,} vs.\ contextual \unittext{organizations}\,$|$\,\unittext{,}); possessives (\unittext{'s}) and quotation marks (\unittext{``}, \unittext{''}) account for the remaining contextual-only units.\looseness=-1

\subsection{Unit and Fixation Visualizations}
\label{app:unit-visualizations}
\begin{figure*}[t]
    \centering
    \includegraphics[width=\textwidth]{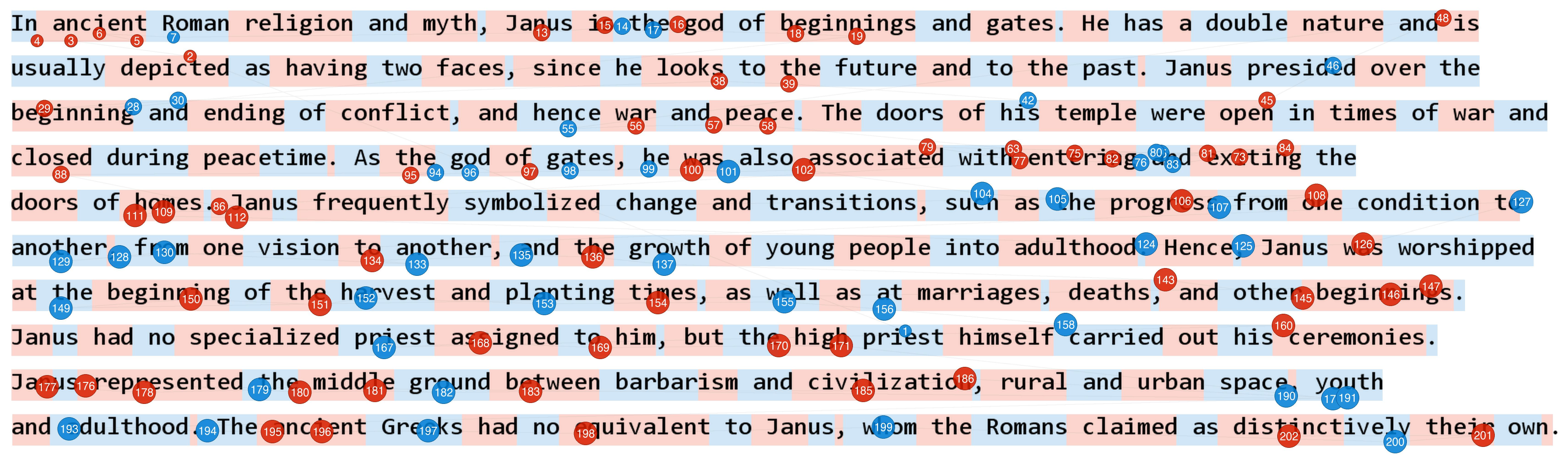}
    \caption{Units and fixations for \textbf{model tokens} (BPE, leading delimiter). Reader~3, Text~1 (MECO English).}
    \label{fig:viz-model-tokens}
\end{figure*}

\begin{figure*}[t]
    \centering
    \includegraphics[width=\textwidth]{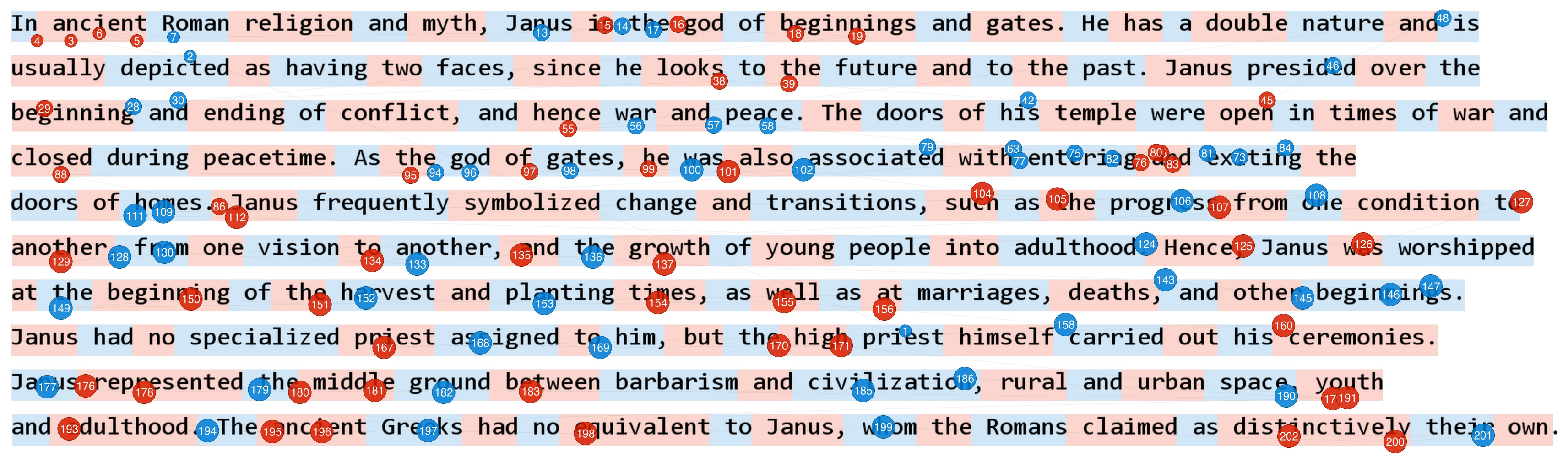}
    \caption{Units and fixations for \textbf{acontextual (leading) words}. Reader~3, Text~1 (MECO English).}
    \label{fig:viz-ws-lead}
\end{figure*}

\begin{figure*}[t]
    \centering
    \includegraphics[width=\textwidth]{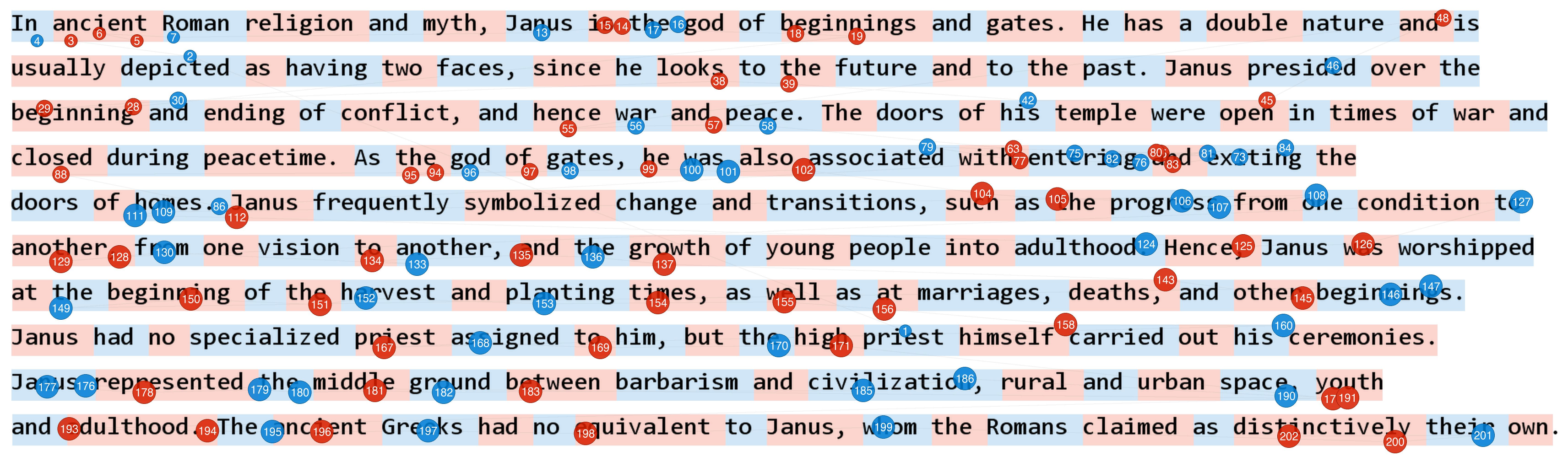}
    \caption{Units and fixations for \textbf{acontextual (trailing) words}. Reader~3, Text~1 (MECO English).}
    \label{fig:viz-ws-trail}
\end{figure*}

\begin{figure*}[t]
    \centering
    \includegraphics[width=\textwidth]{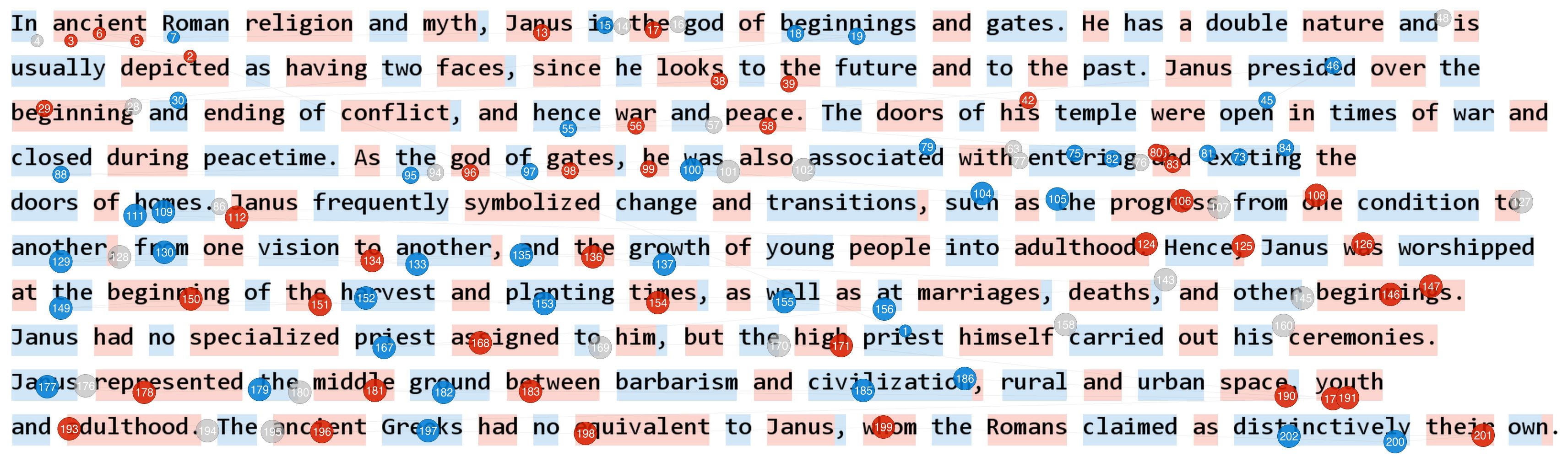}
    \caption{Units and fixations for \textbf{contextual words}. Reader~3, Text~1 (MECO English).}
    \label{fig:viz-ptb}
\end{figure*}

\begin{figure*}[t]
    \centering
    \includegraphics[width=\textwidth]{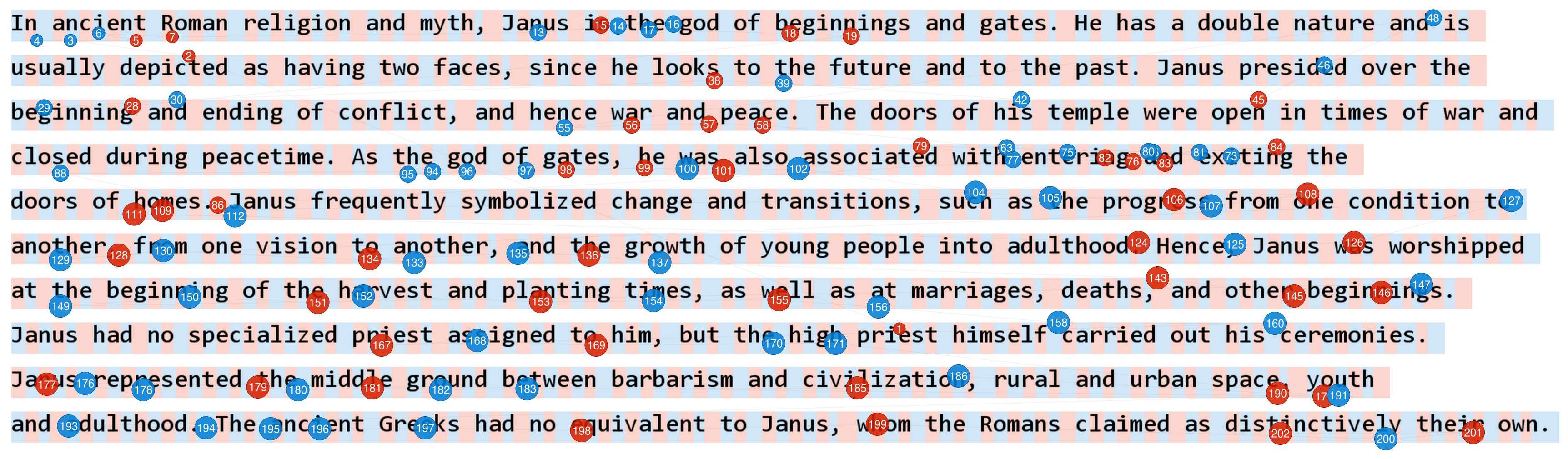}
    \caption{Units and fixations for \textbf{character-level units} (leading delimiter). Reader~3, Text~1 (MECO English).}
    \label{fig:viz-character}
\end{figure*}

\Cref{fig:viz-model-tokens,fig:viz-ws-lead,fig:viz-ws-trail,fig:viz-ptb,fig:viz-character} show the same trial (Reader~3, Text~1 from the MECO English corpus) segmented under the unit inventories used in our experiments, together with the recorded fixation data. Coloured backgrounds mark unit boundaries, and each fixation is coloured by the unit it is attributed to. In the contextual (PTB) inventory, a sentence-final period is split off as its own unit typically only when it is followed by $\eos$; periods that end a sentence mid-trial remain attached to the preceding word.\looseness=-1

\clearpage

\section{Experimental Details}
\label{app:experiments}
To compute surprisal estimates for the experiments in \cref{sec:experiments}, we use the implementation by \citet{snbjarnarson2026transducing} to compose GPT-2 Small\footnote{\href{https://huggingface.co/openai-community/gpt2}{openai-community/gpt2}} \citep{radford2019language} from the \raisebox{-0.2ex}{\huggingface}\kern3pt Hugging Face hub~\citep{wolf-etal-2020-transformers} with the respective transducers described in \cref{sec:units}. To convert token-level models to character-level, we use \raisebox{-0.4ex}{\includegraphics[height=0.9em]{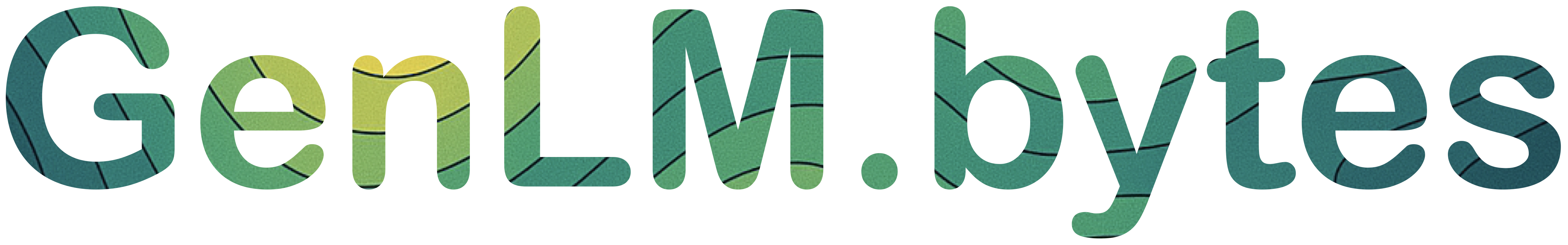}}.\footnote{\href{https://github.com/genlm/genlm-bytes}{genlm-bytes}} To quickly compute next-token/byte distributions, we use {\scriptsize\raisebox{-0.2ex}{\includegraphics[height=1em]{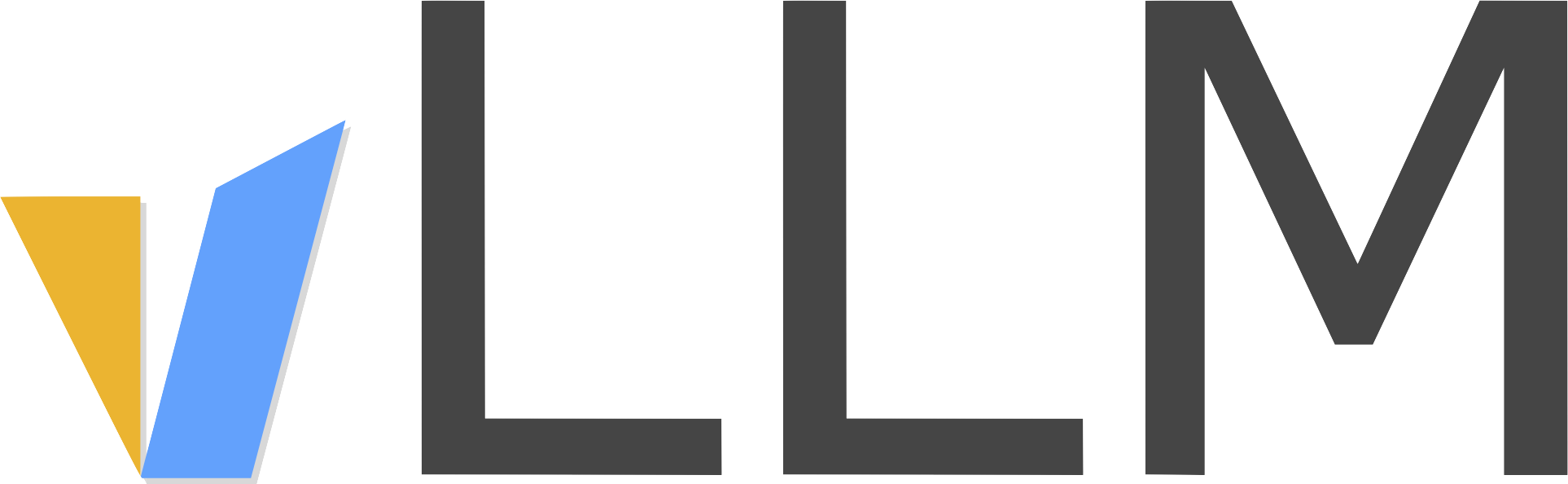}}} \citep{kwon-etal-2023-vllm}.\looseness=-1

\subsection{Computing Surprisal}

Both contextual surprisal and unigram surprisal are computed under the transduced language model $\pTarget = \pSource\circ\transST$, from the next-unit conditional distribution $\prefixU(\unit \mid \utterance) = \prefixTarget(\transH(\unit) \mid \transH(\utterance))$ of \cref{eq:next-unit}. Following $\transH$ as defined in \cref{eq:unit-homomorphism}, each unit's byte extension ends with a $\SEP$ that marks its right boundary. Per-unit conditional probability is therefore the ratio
\begin{equation}
\prefixU(\unit \mid \utterance_{<t}) \;=\; \frac{\prefixTarget\!\big(\transH(\utterance_{<t}) \c \transH(\unit)\big)}{\prefixTarget\!\big(\transH(\utterance_{<t})\big)},
\end{equation}
in which the numerator is a byte-level prefix mass closed off by the $\SEP$ carried in $\transH(\unit)$, making $\transH$ prefix-free (see \cref{fn:prefix-free}). The two quantities (contextual vs.\ unigram surprisal) differ only in how they consume this conditional: contextual surprisal scores the unit that actually occurred at each position, whereas unigram surprisal computes the marginal $\prefixU(\unit) = \mathbb{E}_{\utterance_{<t}}[\prefixU(\unit \mid \utterance_{<t})]$ with respect to contexts $\utterance_{<t}$ sampled from the LM, with $\unit$ held fixed. Both share the hyperparameters listed in \cref{tab:surprisal-params}.

\paragraph{Contextual surprisal.}
For each of the 12 MECO trials we first apply the transducer $\transST$ at the trial level to obtain the transduced string $\tgtStr \in \tgtStrings$, then score its symbols left-to-right. Each step issues one call to the fast next-symbol decomposition (\decomposeNext) algorithm of \citet[\S C.4]{snbjarnarson2026transducing}, which returns the full next-symbol distribution over $\tgtAlphabet$. Consecutive calls are cached, so advancing the context from $\tgtStr_{<t}$ to $\tgtStr_{<t+1}$ extends the cached decomposition by a single symbol rather than recomputing from scratch; low-probability beams are pruned during expansion using the thresholds in \cref{tab:surprisal-params}.

\paragraph{Unigram surprisal.}
Unigram surprisal can be estimated using the same LM that estimates the surprisal of $\unit$ under $\pU$ by marginalizing over contexts \citep{hopton2026unigram}. Because the full marginalization is intractable, we compute a Monte Carlo estimate by drawing $S$ samples from the LM, and for each sample, average the next-unit conditional $\prefixU(\unit \mid \utterance_{<t})$ over every unit position $t$. For the \textbf{GPT-2 token} and \textbf{character} inventories, the unit alphabet coincides with the native output alphabet of the LM, so the conditional $\prefixU(\unit \mid \utterance_{<t})$ is read off directly from the LM. For \textbf{acontextual} and \textbf{contextual} inventories, units are defined through $\transST$ and $\prefixU(\unit \mid \utterance_{<t})$ must be recovered from $\prefixTarget$: We first transduce each sample from $\pSource$ to its target string $\tgtStr^{s} \in \tgtStrings$ and locate the $\SEP$ positions $b_1 < \cdots < b_{K_s}$ that mark unit boundaries. At each boundary $b_k$, we cache the closed prefix mass $z = \log\prefixTarget(\tgtStr^{s}_{\le b_k})$, i.e., the prefix mass through and including the $\SEP$ at position $b_k$, and score every candidate unit $\unit \in \unitAlphabet$ by evaluating $z_\unit = \log\prefixTarget(\tgtStr^{s}_{\le b_k} \c \transH(\unit))$, so that the next-unit conditional is $\prefixU(\unit \mid \utterance_{<t}) = \exp(z_\unit - z)$ exactly, following \cref{eq:next-unit}.

\paragraph{Efficient scoring for the transduced LM.}
Rather than computing the full next-distribution (\decomposeNext), we use the single-symbol scoring routine of \citet{snbjarnarson2026transducing} (introduced there for cross-entropy evaluation), which decomposes only $\tgtStr \cdot \tgtSym$ for a specified target symbol and returns $\log\prefixTarget(\tgtStr \cdot \tgtSym)$ directly, skipping the full-vocabulary expansion and the final normalization. We apply this routine one target symbol at a time along the byte extension of each unit, accumulating log prefix masses and recovering the unit's conditional prefix probability by subtracting the cached boundary probability mass. Since all $|\unitAlphabet|$ unit extensions at a given boundary begin from the same prefix, their decomposition is shared across units; units with common byte prefixes additionally reuse cached partial extensions.

\paragraph{Parameters.}
\cref{tab:surprisal-params} lists the hyperparameters for both computations. The beam-search and transduced-LM parameters are shared; the sampling block applies only to unigram estimation.

\begin{table}[htp]
\centering
\small
\begin{tabular}{@{}lll@{}}
\toprule
\textbf{Parameter} & \textbf{Value} & \textbf{Role} \\
\midrule
\multicolumn{3}{@{}l}{\raisebox{-0.4ex}{\includegraphics[height=0.9em]{emoji/GenLMBytes}} \citep{pmlr-v267-vieira25a}} \\
\midrule
Beam size ($K$) & $5$ & Maximum beam width; keeps the $K$ highest-probability beams. \\
Beam prune threshold & $0.001$ & Drop beams whose probability mass falls below this threshold. \\
\midrule
\multicolumn{3}{@{}l}{Transduced-LM inference \citep{snbjarnarson2026transducing}} \\
\midrule
$\transST$ prune threshold ($\tau$) & $0.005$ & Drop FST paths with probability mass below~$\tau$. \\
Max expand steps & $5$ & Halt expansion of non-universal states after this many steps. \\
Expand stop mass & $0.01$  & Halt expansion once the relative remaining probability mass falls below this value. \\
\midrule
\multicolumn{3}{@{}l}{Sampling for estimating unigram surprisal} \\
\midrule
LM $\pSource$ & GPT-2 Small & Source language model used in all experiments. \\
\# samples $S$ & $500$ & Number of samples drawn from $\pSource$ in the unigram estimator. \\
Max length & $50$ & Maximum number of tokens per sample. \\
Batch size & $64$ & Batch size used when sampling from $\pSource$. \\
\bottomrule
\end{tabular}
\caption{Hyperparameters used to estimate contextual and unigram surprisal. The beam-search parameters apply to both estimators; the sampling parameters apply only to the unigram estimator.}
\label{tab:surprisal-params}
\end{table}

\subsection{GPU Usage \& Runtime}
All experiments were run on NVIDIA GeForce RTX 4090 GPUs and RTX 3090 GPUs, each with 24\,GB of GPU memory. \Cref{tab:throughput-surprisal} reports scoring throughput for contextual surprisal. The acontextual FSTs process approximately 200 target symbols per second; at this rate, scoring the 12 MECO English trials takes approximately one minute on a single GPU. The contextual FST is more than an order of magnitude slower, due to its large number of states and arcs (see \cref{tab:transducer-sizes}) and the fact that many of its states are not universal, requiring additional computation to traverse the FST until hitting a universal state; see \cref{app:transducer-sizes} for a brief discussion and \citet{snbjarnarson2026transducing} for a detailed discussion on universality.

\begin{table}[htp]
\centering
\small
\caption{Contextual surprisal scoring throughput per transducer on GPT-2 Small, aggregated over the 12 MECO English trials. ``Symbols'' is the total number of $\tgtAlphabet$-symbols scored across all trials; ``Syms/s'' is the corresponding throughput. The character-level model is omitted because character surprisal is obtained via \citet{pmlr-v267-vieira25a} rather than through an FST composition.}
\label{tab:throughput-surprisal}
\begin{tabular}{lrrr}
\toprule
Transducer & Symbols & Time (s) & Syms/s \\
\midrule
Acontextual (leading) & 15{,}309 &   72.0 & 212.8 \\
Acontextual (trailing) & 15{,}309 &   75.2 & 203.6 \\
Contextual & 13{,}407 & 1117.7 &  12.0 \\
\bottomrule
\end{tabular}
\end{table}

Unigram estimation is considerably more expensive than contextual surprisal, because every $\SEP$-boundary of every sample requires an extension of the cached prefix mass for every unit in $\unitAlphabet$. On GPT-2, typical per-sample scoring of a 211-byte sample takes on the order of $20$~seconds for acontextual (leading), $60$~seconds for acontextual (trailing), and $8$~minutes for the contextual transducer; the relative ordering mirrors \cref{tab:throughput-surprisal} because the same FST-decomposition step is the bottleneck in both pipelines. Run sequentially, at 500 samples per unit inventory, this translates to roughly 3 hours (acontextual leading), 8 hours (acontextual trailing), and 2--3 days (contextual) of sequential single-GPU compute, so unigram estimation in practice requires parallelization. Since the outer sample loop runs independently across $s$, we chunk the 500-sample runs into independent jobs. Some chunks hit the per-job wall-clock limit before completing, so the final number of successfully scored samples per inventory is 498 (characters and GPT-2 tokens), 496 (acontextual leading), 497 (acontextual trailing), and 470 (contextual); at these sample sizes, the per-unit probabilities are close to converged, and additional samples shift the estimates only marginally.

\clearpage
\section{GAMM Specification}
\label{sec:gamm-spec}
We model log reading time as a generalized additive mixed model \citep{wood-2017-gam} fitted with \texttt{bam()} from \texttt{mgcv} in R, using fast restricted maximum likelihood (\texttt{method="fREML"}, \texttt{discrete=TRUE}). Each continuous predictor enters the log-mean as a cubic regression spline with up to six basis functions (\texttt{bs="cr"}, \texttt{k=6}); the model further includes a random intercept for participant and by-participant random slopes for every continuous predictor.

\vspace{1em}
\noindent\begin{minipage}[t]{0.48\textwidth}
\paragraph{Baseline model ($\widetilde{\varphi}$).}
\begin{small}
\begin{verbatim}
log(rt) ~ s(length, bs="cr", k=6)
  + s(length_prev, bs="cr", k=6)
  + s(length_prev2, bs="cr", k=6)
  + s(unigram_surprisal, bs="cr", k=6)
  + s(unigram_surprisal_prev, bs="cr", k=6)
  + s(unigram_surprisal_prev2, bs="cr", k=6)
  + s(length, participant, bs="re")
  + s(length_prev, participant, bs="re")
  + s(length_prev2, participant, bs="re")
  + s(unigram_surprisal, participant,
      bs="re")
  + s(unigram_surprisal_prev, participant,
      bs="re")
  + s(unigram_surprisal_prev2, participant,
      bs="re")
  + s(participant, bs="re")
\end{verbatim}
\end{small}
\end{minipage}\hfill
\begin{minipage}[t]{0.48\textwidth}
\paragraph{Target model ($\varphi$).} The target model adds contextual surprisal and its two spillover lags to the baseline:
\begin{small}
\begin{verbatim}
log(rt) ~ ... [baseline terms] ...
  + s(surprisal, bs="cr", k=6)
  + s(surprisal_prev, bs="cr", k=6)
  + s(surprisal_prev2, bs="cr", k=6)
  + s(surprisal, participant, bs="re")
  + s(surprisal_prev, participant, bs="re")
  + s(surprisal_prev2, participant, bs="re")
\end{verbatim}
\end{small}
\end{minipage}
\vspace{1em}

\noindent Here \texttt{s(x, bs="cr", k=6)} denotes a cubic regression spline with 6 basis functions; \texttt{length} is unit length in characters (whitespace-inclusive: for the acontextual and model-tokens inventories, a unit's length includes any leading or trailing whitespace attributed to it by the transducer, so a mid-sentence word is one character longer than its raw spelling); \texttt{unigram\_surprisal} is unigram surprisal (see \cref{sec:baseline}); \texttt{surprisal} is contextual surprisal from the language model; \texttt{\_prev} and \texttt{\_prev2} denote spillover from the first and second preceding unit, respectively. Each predictor enters as both a population-level smooth and a by-participant random slope \texttt{s(x, participant, bs="re")}; a random intercept for participants is included via \texttt{s(participant, bs="re")}.
For the character inventory, unit length is constant (every unit is a single character), so the \texttt{length}, \texttt{length\_prev}, and \texttt{length\_prev2} smooths are omitted from both the baseline and target formulas; the remaining terms and the random-effects structure are unchanged.

\paragraph{Paired permutation test.}
Significance of $\dll$ is assessed by a one-sided paired permutation test on the per-observation held-out log-likelihood differences between the target and baseline models, with $B = 1000$ sign-flip permutations.

\section{Additional Results}
\cref{tab:gamm_all} reports detailed GAMM results for each reading-time measure: mean per-observation held-out log-likelihood for the baseline ($\llb$) and target ($\lltgt$) models, along with the improvement $\dll$ and 95\% trial-level bootstrap CIs. Note that absolute log-likelihoods are not comparable across unit inventories because the number and granularity of observations differ. \cref{tab:units_stats_gamm} reports the number of observations entering the regression for each inventory, after the additional exclusions specific to the GAMM input.

\begin{figure*}[h]
    \centering
    \includegraphics[width=\textwidth]{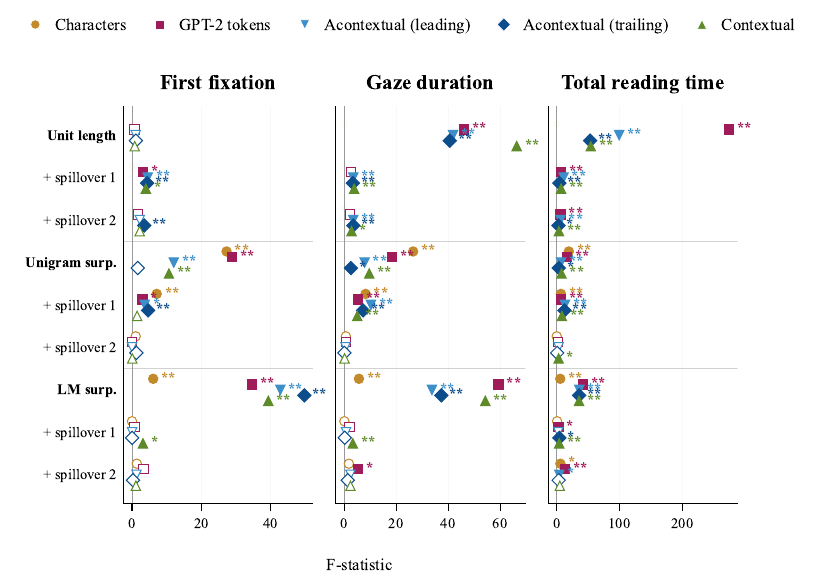}
    \captionof{figure}{Approximate F-statistics for fixed-effect smooth terms from the full-data GAMM fit, grouped by predictor type. Within each group, rows correspond to the current unit, spillover~1, and spillover~2. Filled markers indicate significance ($^{*}$\,$\significance < 0.05$; $^{**}$\,$\significance < 0.01$).}
    \label{fig:gamm_coefficients}
\end{figure*}

\begin{table}[htp]
\centering
\small
\begin{tabular}{@{}l r@{}}
\toprule
\textbf{Inventory} & $n_{\text{GAMM}}$ \\
\midrule
GPT-2 tokens         & 40{,}589 \\
Acontextual (leading)  & 39{,}290 \\
Acontextual (trailing) & 39{,}767 \\
Contextual           & 33{,}472 \\
Characters           & 48{,}834 \\
\bottomrule
\end{tabular}
\caption{Final number of observations entering the GAMM per unit inventory, starting from $n_{\text{lag}}$ (\cref{tab:units_stats}) after additionally excluding observations with missing unigram surprisal (Contextual only: 42 obs for $\srcfont{``}$ and $\srcfont{''}$, whose candidate-scoring paths fall below the beam prune thresholds in \cref{tab:surprisal-params} at every sampled context) or zero surprisal (Characters only: 4{,}780 sub-token byte positions where the byte distribution is deterministic under BPE).}
\label{tab:units_stats_gamm}
\end{table}
\clearpage
\begin{table}[h]
\centering
\caption{GAMM results across reading-time measures. $\llb$/$\lltgt$: mean per-obs.\ held-out log-likelihood of the baseline/target; $\dll$: improvement ($\times 10^{-3}$ nats) with 95\% trial-level bootstrap CI in brackets. Significance via paired permutation test (see \cref{sec:gamm-spec}): $^{*}$\,$\significance < 0.05$; $^{**}$\,$\significance < 0.01$.}
\label{tab:gamm_all}
\resizebox{\textwidth}{!}{%
\begin{tabular}{@{}l rrrr rrrr rrrr@{}}
\toprule
 & \multicolumn{4}{c}{\textit{First fixation}} & \multicolumn{4}{c}{\textit{Gaze duration}} & \multicolumn{4}{c}{\textit{Total reading time}} \\
\cmidrule(lr){2-5}\cmidrule(lr){6-9}\cmidrule(lr){10-13}
Inventory & $\llb$ & $\lltgt$ & $\dll$ & $\significance$ & $\llb$ & $\lltgt$ & $\dll$ & $\significance$ & $\llb$ & $\lltgt$ & $\dll$ & $\significance$ \\
\midrule
Characters
  & $-0.4958$ & $-0.4957$ & \shortstack[r]{$0.11$\phantom{$^{**}$}\\[-2pt]{\tiny$[-0.16,\, 0.40]$}} & $0.145$
  & $-0.5008$ & $-0.5008$ & \shortstack[r]{$0.09$\phantom{$^{**}$}\\[-2pt]{\tiny$[-0.15,\, 0.35]$}} & $0.185$
  & $-0.5662$ & $-0.5661$ & \shortstack[r]{$0.10$\phantom{$^{**}$}\\[-2pt]{\tiny$[-0.22,\, 0.45]$}} & $0.171$ \\[6pt]
GPT-2 tokens
  & $-0.4725$ & $-0.4720$ & \shortstack[r]{$0.55^{*}$\\[-2pt]{\tiny$[-0.37,\, 1.46]$}} & $0.013$
  & $-0.5958$ & $-0.5943$ & \shortstack[r]{$1.52^{**}$\\[-2pt]{\tiny$[0.09,\, 2.85]$}} & $<\!0.001$
  & $-0.7562$ & $-0.7537$ & \shortstack[r]{$2.56^{**}$\\[-2pt]{\tiny$[0.07,\, 4.98]$}} & $<\!0.001$ \\[6pt]
Acontextual (leading)
  & $-0.4699$ & $-0.4697$ & \shortstack[r]{$0.28$\phantom{$^{**}$}\\[-2pt]{\tiny$[-0.69,\, 1.16]$}} & $0.065$
  & $-0.6102$ & $-0.6088$ & \shortstack[r]{$1.41^{**}$\\[-2pt]{\tiny$[-0.14,\, 2.81]$}} & $<\!0.001$
  & $-0.7681$ & $-0.7651$ & \shortstack[r]{$3.00^{**}$\\[-2pt]{\tiny$[0.48,\, 5.34]$}} & $<\!0.001$ \\[6pt]
Acontextual (trailing)
  & $-0.4692$ & $-0.4686$ & \shortstack[r]{$0.63^{**}$\\[-2pt]{\tiny$[-0.67,\, 1.76]$}} & $0.004$
  & $-0.6061$ & $-0.6044$ & \shortstack[r]{$1.68^{**}$\\[-2pt]{\tiny$[-0.12,\, 3.17]$}} & $<\!0.001$
  & $-0.7669$ & $-0.7640$ & \shortstack[r]{$2.91^{**}$\\[-2pt]{\tiny$[-0.16,\, 5.35]$}} & $<\!0.001$ \\[6pt]
Contextual
  & $-0.4745$ & $-0.4737$ & \shortstack[r]{$0.81^{**}$\\[-2pt]{\tiny$[-0.47,\, 1.94]$}} & $0.003$
  & $-0.5984$ & $-0.5962$ & \shortstack[r]{$2.13^{**}$\\[-2pt]{\tiny$[0.43,\, 3.73]$}} & $<\!0.001$
  & $-0.7409$ & $-0.7376$ & \shortstack[r]{$3.24^{**}$\\[-2pt]{\tiny$[-0.16,\, 6.32]$}} & $<\!0.001$ \\
\bottomrule
\end{tabular}%
}
\end{table}
\cref{fig:gamm_coefficients} shows the approximate F-statistics for all fixed-effect smooth terms in the full-data GAMM fit, broken down by predictor group and reading-time measure. We observe two patterns: First, the current-unit surprisal smooth is significant for every unit inventory and every reading-time measure. Second, length and spillover predictors contribute primarily to the word-like inventories (tokens, acontextual, contextual) and to the later measures (gaze duration and total reading time); at the character level, the length smooth is absent because all units share the same length.

\end{document}